\documentclass[letterpaper,twocolumn,10pt]{article}
\usepackage{usenix}
\usepackage{titling}
\usepackage{tikz}
\usepackage{multirow} 
\usepackage{pifont}
\usepackage{amsmath}
\usepackage{amssymb}
\usepackage{url}
\usepackage{subcaption}
\usepackage{graphicx}
\usepackage{booktabs}
\usepackage{algpseudocode}
\usepackage[linesnumbered,ruled,vlined]{algorithm2e}
\usepackage{tabularx}
\usepackage{authblk}
\usepackage{tablefootnote}
\usepackage{float}

\usepackage{enumitem} 
\usepackage{pifont}
\usepackage{placeins}
\newcommand{\mixed}{\ding{51}\kern-0.62em\ding{55}} 

\usepackage{tikz}
\newcommand*\circled[1]{%
  \scalebox{0.78}{\begin{tikzpicture}[baseline=-3pt]
    \node[draw,circle,inner sep=0.5pt, fill=black] {\textcolor{white}{\textsf{\textbf{#1}}}};
  \end{tikzpicture}}}

 \usepackage{etoolbox}
 
 \makeatletter
 \patchcmd{\maketitle}
 	{\@maketitle}
 	{\vspace{-7em}\@maketitle\vspace{-5em}}
 	{}
 	{}
 \makeatother

\algdef{SE}[LOOP]{Loop}{EndLoop}[1]{\textbf{loop} #1}{}

\begin{document}
\newcommand{\gemm}{MeshGEMM\xspace}
\newcommand{\gemv}{MeshGEMV\xspace}
\newcommand{\sys}{WaferLLM\xspace}


\newcommand{\para}[1]{\noindent\textbf{#1}}

\newcommand{\tinyskip}{\vspace{3pt}}
\newcommand{\mypar}[1]{\tinyskip\noindent\textbf{#1.}\xspace}
\newcommand{\myitem}[1]{\item\textbf{#1.}\xspace}

\newcommand*\myc[1]{%
\scalebox{0.78}{\begin{tikzpicture}[baseline=-3pt]
  \node[draw,circle,inner sep=0.5pt, fill=black] {\textcolor{white}{\textsf{\textbf{#1}}}};
\end{tikzpicture}}}


\makeatletter
\DeclareRobustCommand\onedot{\futurelet\@let@token\@onedot}
\def\@onedot{\ifx\@let@token.\else.\null\fi\xspace}

\def\eg{\emph{e.g}\onedot} \def\Eg{\emph{E.g}\onedot}
\def\ie{\emph{i.e}\onedot} \def\Ie{\emph{I.e}\onedot}
\def\cf{\emph{c.f}\onedot} \def\Cf{\emph{C.f}\onedot}
\def\etc{\emph{etc}\onedot} \def\vs{\emph{vs}\onedot}
\def\wrt{w.r.t\onedot} \def\dof{d.o.f\onedot}
\def\etal{\emph{et al}\onedot}
\makeatother
\newcommand{\Sum}[3]{\sum\limits_{#1}^{#2}{#3}}
\newcommand{\lr}[3]{\left #1 {#3} \right #2}
\newcommand{\E}[2]{E[#1,#2]}

\newenvironment{tightlist}{
\begin{list}{$\bullet$}{
    \setlength{\topsep}{.1em}
    \setlength{\partopsep}{0in}
    \setlength{\parskip}{0in}
    \setlength{\itemsep}{0in}
    \setlength{\parsep}{0in}
    \setlength{\leftmargin}{1em}
    \setlength{\rightmargin}{0in}
    \setlength{\itemindent}{0in}
}}
{\end{list}}


\title{\Large \bf \sys: Large Language Model Inference at Wafer Scale}
\date{}



\author{
{\rm Congjie He$^1$ \hspace{0.05cm} Yeqi Huang$^1$ \hspace{0.05cm} Pei Mu$^1$ \hspace{0.05cm} Ziming Miao$^2$ \hspace{0.05cm} Jilong Xue$^2$ \hspace{0.05cm} Lingxiao Ma$^2$ \hspace{0.05cm} Fan Yang$^2$ \hspace{0.05cm} Luo Mai$^1$} \\ \vspace{-0.1cm}
{\it $^1$University of Edinburgh \quad $^2$Microsoft Research \vspace{-0.5cm}}
}
\maketitle

\begin{abstract}

Emerging AI accelerators increasingly adopt wafer-scale manufacturing technologies, integrating hundreds of thousands of AI cores in a mesh architecture with large distributed on-chip memory (tens of GB in total) and ultra-high on-chip memory bandwidth (tens of PB/s). However, current LLM inference systems, optimized for shared memory architectures like GPUs, fail to exploit these accelerators fully.

We introduce \sys, the first wafer-scale LLM inference system. \sys is guided by a novel PLMR model (pronounced as "Plummer") that captures the unique hardware characteristics of wafer-scale architectures. Leveraging this model, \sys pioneers wafer-scale LLM parallelism, optimizing the utilization of hundreds of thousands of on-chip cores. It also introduces \gemm and \gemv, the first GEMM and GEMV implementations designed to scale effectively on wafer-scale accelerators.

Evaluations show that \sys achieves up to 200$\times$ higher accelerator utilization than state-of-the-art methods. Leveraging a wafer-scale accelerator (Cerebras WSE2), \sys delivers GEMV operations 606$\times$ faster and 16$\times$ more energy-efficient than on an NVIDIA A100 GPU. For full LLM inference, \sys achieves 10–20$\times$ speedups over A100 GPU clusters running SGLang and vLLM. These advantages are expected to grow as wafer-scale AI models, software, and hardware continue to mature. \sys is open-sourced at \href{https://github.com/MeshInfra/WaferLLM}{https://github.com/MeshInfra/WaferLLM}.

\end{abstract}

\section{Introduction}
\vspace{-0.3cm}

Large Language Model (LLM) inference is a rapidly growing workload. It has two phases~\cite{flashdecoding++}: (i) the \textit{prefill phase}, which processes input tokens (the prompt) and spends most of its cycles on General Matrix Multiply (GEMM); and (ii) the \textit{decode phase}, which generates tokens one by one in an autoregressive manner, primarily performing General Matrix-Vector Product (GEMV). Decode requires repeatedly loading the entire LLM model into on-chip memory, with GEMV dominating its cycles. Since LLMs generate many tokens, especially in the test-time scaling scenario, such as the OpenAI-o1/o3~\cite{openaio1,openaio3} and DeepSeek-R1~\cite{deepseekr1}, inference is constrained by GEMV latency, making it inherently memory-bandwidth-bound.

To address memory bandwidth bottlenecks, AI accelerators are increasingly adopting system-on-wafer integration~\cite{tsmc-advanced-packaging}. This approach scales chip area to a full wafer, up to 100$\times$ larger than a typical GPU die, enabling significantly more on-chip cores, memory, and bandwidth. Examples include Cerebras WSE~\cite{wse} and upcoming Tesla Dojo~\cite{dojo}. The Cerebras WSE-2, for instance, integrates 850,000 cores with 40GB of on-chip memory, 1,000$\times$ more than GPUs, and provides 22PB/s memory bandwidth, 7,000$\times$ higher than GPUs. TSMC anticipates widespread adoption of system-on-wafer integration, citing advantages in performance, energy-efficient die-to-die communication, and cost reduction. IEEE similarly forecasts a surge in wafer-scale computing by 2027~\cite{tsmc-advanced-packaging}. Wafer-scale accelerators are already seeing real-world deployment, particularly in model serving. In February 2025, Mixtral and Perplexity adopted wafer-scale chips, achieving cost parity with GPUs in terms of tokens per dollar~\cite{perplexity,mistra}. G42 now operates data centers fully outfitted with wafer-scale accelerators, and Cerebras has secured major commercial contracts~\cite{g42}.

Unlocking the potential of wafer-scale accelerators is challenging because current LLM systems rely on \emph{shared memory architectures} typical of GPUs and TPUs. Wafer-scale accelerators, however, adopt \emph{network-on-chip} (NoC) designs that interconnect millions of cores with local memory in a \emph{massive-scale, mesh-based memory architecture}. This architecture far exceeds the scale of on-chip crossbars (e.g., one-hop NUMA such as GraphCore IPU), multi-socket NUMA~\cite{amd2023numa}, and high-density AI clusters (hundreds of GPUs per pod)~\cite{tpu}. Without fully addressing this fundamental shift in memory architecture, directly applying designs from state-of-the-art systems like T10~\cite{t10} and Ladder~\cite{ladder} to wafer-scale devices often results in extremely poor performance.

To address these challenges, we propose a \emph{device model} that captures the critical hardware properties of wafer-scale accelerators, highlighting key differences from shared-memory devices. This model enables us to evaluate current LLM inference design principles, identify non-compliant areas, and pinpoint where new approaches are required. Guided by this model, we can achieve an ambitious system design: \emph{running complete LLM inference on a single chip}, minimizing costly off-chip communication and maximizing on-chip memory bandwidth utilization.

The above idea motivates \sys, the first wafer-scale LLM inference system, yielding several contributions:

\mypar{(1) Device model for wafer-scale accelerators}
We propose the PLMR model\footnote{PLMR model can be pronounced as “Plummer” }, which captures the key hardware properties of wafer-scale accelerators:
(i) Massive \textbf{P}arallel cores (P): Millions of cores can be integrated on a large wafer, requiring systems to effectively partition LLMs and their operations.
(ii) Highly non-uniform memory access \textbf{L}atency (L): Inter-core data access exhibits significant variation, with latency differences up to 1,000$\times$, necessitating the system to mitigate this.
(iii) Constrained per-core local \textbf{M}emory (M): Each core has limited memory (tens of KBs to several MBs), requiring efficient memory usage.
(iv) Limited hardware-assisted \textbf{R}outing (R): The NoC routing hardware is constrained by the area size per core and can only support a limited number of routing paths, e.g., less than $2^5$ on Cerebras WSE-2.

\mypar{(2) Wafer-scale LLM parallelism}
We propose an effective, PLMR-compliant LLM parallelism policy for wafer-scale accelerators. 
In the prefill phase, we design fine-grained partitioning to achieve million-core parallelism. For the decode phase, where tensor dimensions are insufficient for partitioning, we design fine-grained replication to enable parallelism with minimal communication costs. As a result, \sys achieves larger-scale and finer-grained parallelism (satisfying P in PLMR) than GPU-based approaches. Additionally, we replace conventional GPU-based GEMM and GEMV operators 
with new algorithm designed for the PLMR model (satisfying L, M, and R) and propose tensor placement strategies that eliminate matrix transpositions, which are costly with a mesh NoC (satisfying L). 

We also designed a scalable KV-cache management method for wafer-scale devices. This approach features a novel KV cache shift method to ensure balanced core usage (satisfying P and M), avoiding skewed utilization of cores caused by KV cache concatenation methods common on GPUs.

\mypar{(3) Wafer-scale GEMM}
We propose \gemm, a scalable GEMM algorithm for wafer-scale devices, accelerating the prefill phase.
Unlike conventional distributed GEMM algorithms, \gemm achieves full PLMR compliance by leveraging two key operations: \emph{cyclic shifting} and \emph{interleaving}. Cyclic shifting ensures algorithm correctness while maintaining bounded usage of local memory (satisfying M). The interleaving operation minimizes communication latency in the mesh NoC, effectively reducing the overhead of highly non-uniform memory latency (satisfying L) and routing resources (satisfying R).

\mypar{(4) Wafer-scale GEMV} 
We propose \gemv, a scalable GEMV algorithm for wafer-scale devices, accelerating the decode phase. Unlike existing GEMV implementations, \gemv uses a novel \emph{K-tree allreduce} algorithm to aggregate local GEMV results across massive cores. This algorithm ensures routing resource usage meets the hardware's limitation (satisfying R) and reduces communication latency (satisfying L).

\vspace{0.2cm}
\noindent
We implemented \sys on the Cerebras WSE engine using approximately 7,000 lines of CSL (a C-like programming language) for LLM parallelism, \gemm, and \gemv, and 2,000 lines of Python for loading LLM checkpoints, launching inference, and execution parallelism policy.

We conducted end-to-end LLM inference experiments with various models, including full LLaMA3-8B and LLaMA2-13B, as well as subsets of layers of CodeLLaMA-34B, and QWen2-72B. By combining wafer-scale LLM parallelism, GEMM and GEMV, \sys outperforms state-of-the-art (SOTA) systems:
(i) 100-200$\times$ faster than T10~\cite{t10}, the SOTA system for massive cores with a distributed on-chip memory architecture, and
(ii) 200-400$\times$ faster than Ladder~\cite{ladder}, the SOTA system for shared-memory architectures.

Micro-benchmarks further show that \gemm is 2-3$\times$ faster than SUMMA~\cite{summa}, the default optimized GEMM for Cerebras WSE, and Cannon~\cite{cannon}, the SOTA GEMM for supercomputers with large-scale mesh architectures. \gemv achieves 4-8$\times$ speedups over Cerebras’s optimized GEMV, and 606$\times$ than a single A100 GPU. Additionally, \sys’s cache shift method is up to 400$\times$ more scalable than the KV cache SOTA on GPUs, such as PagedAttention~\cite{vllm}.

Combined, \sys (on Cerebras WSE-2) outperforms SGLang (on single A100) by 30-40$\times$. Compared to the optimal performance of SGLang on A100 multi-GPUs connected with NVLink and RDMA, \sys delivers a 10-20$\times$ faster end-to-end speed and is 2.5$\times$ more energy-efficient. The reduced gains from GEMV to LLM are due to current limitations in software, hardware, and existing LLM model designs. We anticipate stronger performance as wafer-scale AI computing matures and these limitations are addressed.

\section{Background and Motivation}\label{sec:motivation}

\subsection{LLM inference and its key constraint}

\begin{figure}[t!]
    \centering
    \includegraphics[width=1\linewidth]{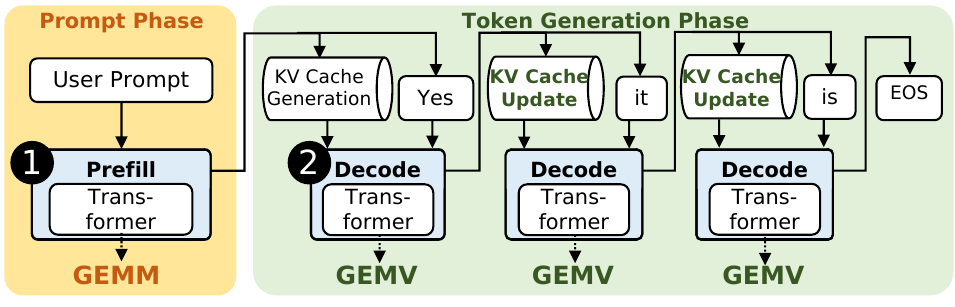}
    \vspace{-0.3cm}
    \caption{Key components in LLM inference}
    \label{fig:llm-inference-overview}
    \vspace{-0.3cm}
\end{figure}

An LLM inference system typically performs auto-regressive token-by-token generation, as illustrated in Figure~\ref{fig:llm-inference-overview}. The model comprises multiple transformer layers, dominated by self-attention and feedforward blocks. Inference operates in two phases: prefill and decode. The total cycles of the prefill phase are dominated by GEMM operations (shown by \circled{1}). While the total cycles of the decode phase are dominated by GEMV operations (shown by \circled{2}).

LLM inference is memory bandwidth-bound. Model weights (10-1,000 GB) are fetched repeatedly from external memory during inference, as GPUs typically have only ~100 MB of on-chip memory. For per request, generating thousands of tokens per second demands hundreds of TB/s bandwidth, far exceeding the capabilities of HBM (high bandwidth memory) on current GPUs. 

While tensor parallelism across GPUs can increase bandwidth, mitigating communication bottlenecks in a large GPU cluster remains challenging. Also, adding GPUs improves throughput for concurrent queries but does not reduce time per output token (TPOT), as each query is still memory bandwidth-limited.

\vspace{-3mm}
\subsection{Reasons for wafer-scale accelerators}
\vspace{-1mm}

\begin{table}[t]
\resizebox{\linewidth}{!}{%
\begin{tabular}{|c|c|c|}
\hline
 & System-on-Die & System-on-Wafer \\ \hline
Area (mm$^2$) & Up to 858 & Up to 73062 \\ \hline
\# Transistors (TSMC n3) & \textasciitilde 1s Tillion & \textasciitilde 10s Trillion \\ \hline
\# Cores & 1,000s-10,000s & 100,000s-1,000,000s \\ \hline
On-Chip Memory & 10s-100s MB & \textasciitilde 10s GB \\ \hline
Memory Bandwidth & 1s TB/s & \textasciitilde 10s PBs/s \\ \hline
Attached HBM & \textasciitilde 10s-100s GB & 10s TB (via TSMC SoW) \cite{tsmc-advanced-packaging} \\ \hline
\begin{tabular}[c]{@{}c@{}}Die-to-Die \\ Bandwidth (TB/s)\end{tabular} & \textasciitilde 1s-10s (via off-chip) & \textasciitilde 10s-100s (via on-chip) \\ \hline
\begin{tabular}[c]{@{}c@{}}Die-to-Die\\ Latency (ns)\end{tabular} & \textasciitilde 100s (via off-chip) & \textasciitilde 1s (via on-chip) \\ \hline
\begin{tabular}[c]{@{}c@{}}Die-to-Die\\ Power (pJ/bit)\end{tabular} & \textasciitilde 10s (via off-chip) & \textasciitilde 0.1s (via on-chip) \\ \hline
\end{tabular}%
}
    \vspace{-2mm}
\caption{System-on-Die vs. System-on-Wafer}
    \vspace{-3mm}
\label{tab1:die-scale-vs-wafer-scale}
\end{table}

To increase memory bandwidth, accelerator designers are increasingly adopting system-on-wafer integration~\cite{tsmc-advanced-packaging} for several reasons:

\mypar{Performance advantages} System-on-wafer technology allows trillions of transistors to be integrated into a single wafer-scale chip, up to 100$\times$ more than a typical GPU die, shown in Table~\ref{tab1:die-scale-vs-wafer-scale}. This enables millions of AI-optimized cores, providing tens of GBs of on-chip memory and up to tens of PB/s memory bandwidth, 1,000$\times$ higher than a standard GPU’s several TB/s. Future wafer-scale chips can also attach 40-80$\times$ more HBM chips to their edge compared to a standard die~\cite{tsmc-advanced-packaging}.

\mypar{Integration efficiency} System-on-wafer excels at integrating massive parallel cores, with wafer-based die-to-die connections offering up to 10$\times$ more bandwidth per unit area, 100-300$\times$ latency benefit, and nearly 100$\times$ better power efficiency per bit than conventional PCB-based chip-to-chip interconnection~\cite{wse} (e.g., NVIDIA NVLink, PCIe), shown in Table~\ref{tab1:die-scale-vs-wafer-scale}. As noted earlier, LLM inference is primarily constrained by memory bandwidth due to intensive data access. In distributed settings, inter-chip communication overhead from remote memory access, especially during decode-phase GEMV, limits scalability. Thus, system-on-wafer integration offers lower-latency and high-bandwidth interconnects and improved efficiency over conventional chip-to-chip links.

\mypar{Lower cost} Wafer-scale integration can lower the manufacturing cost, since a significant fraction of the cost of fabrication (typically 30-50\%) is related to testing and packaging the individual chips~\cite{wiki:waferscale}. Additionally, wafer-scale integration has made notable progress in yield improvement. Companies such as TSMC are also developing techniques to integrate fully tested dies on a single wafer, further enhancing yield.

\vspace{-3mm}
\subsection{Challenges for wafer-scale LLM inference}
\vspace{-1mm}

\begin{figure}[t]
    \centering
    \includegraphics[width=1\linewidth]{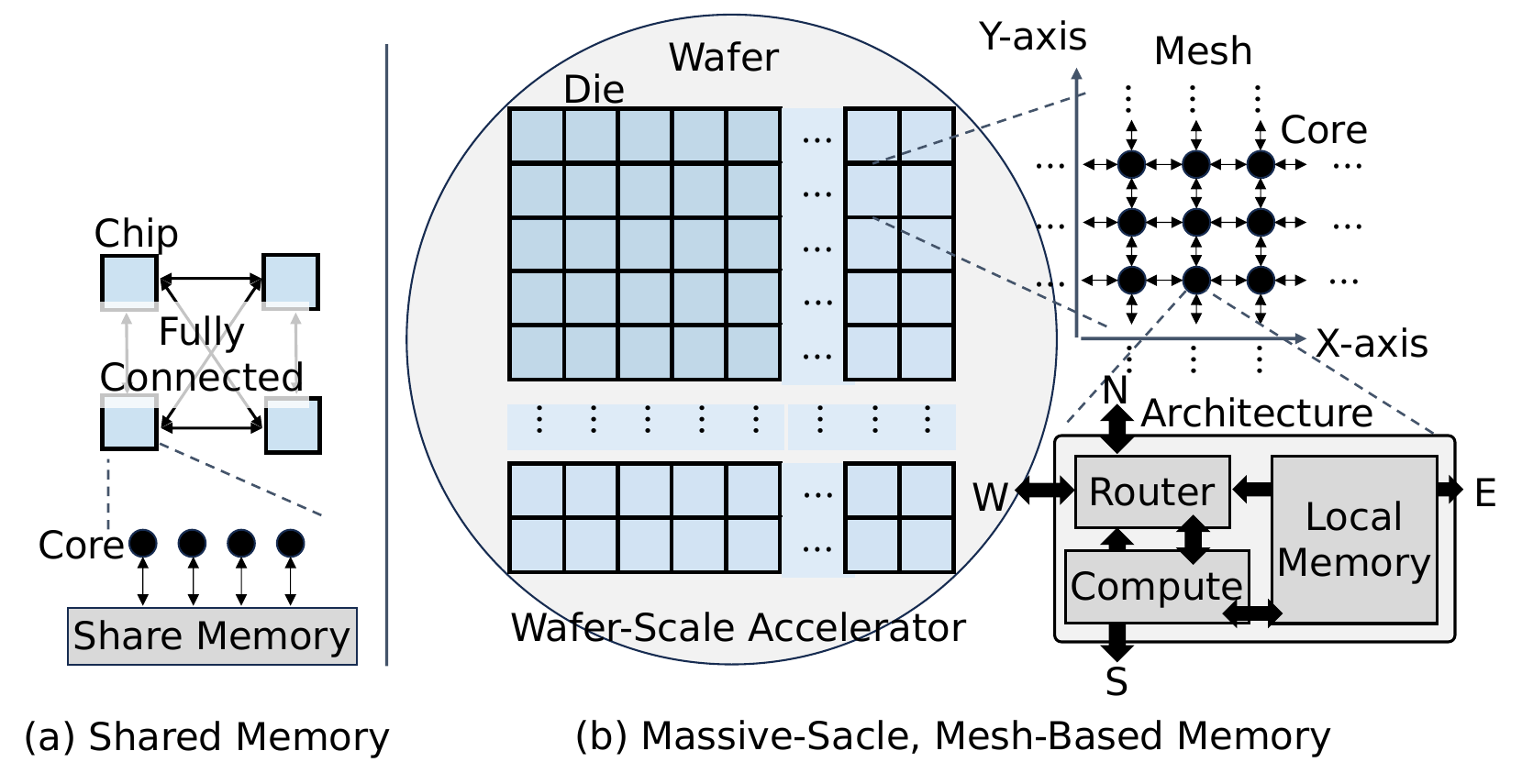}
    \vspace{-3mm}
    \caption{Massive-scale mesh-based memory architecture}
    \vspace{-3mm}
    \label{fig:mesh-architecture}
\end{figure}

The key challenge in leveraging wafer-scale accelerators for LLM inference is their shift to a distributed, non-uniform memory architecture on a single chip. Current LLM systems are optimized for shared memory (single chip) or fully connected architectures (e.g., GPU pods), as shown in Figure~\ref{fig:mesh-architecture}(a). However, as on-chip memory size grows, these architectures face exponential manufacturing costs and performance degradation, driving the need for a distributed on-chip architecture.

AI accelerator designers predominantly use a \textbf{mesh-like network-on-chip (NoC)} to connect \textbf{massive cores} (ranging from hundreds of thousands to millions), as shown in Figure~\ref{fig:mesh-architecture}(b). The mesh topology is favored for its efficiency in core arrangement, enabling effective cooling~\cite{cooling}, power delivery~\cite{powerdelivery}, and cost-efficient wiring~\cite{chip-cost-1, chip-cost-2}, with each core communicating only with nearby neighbors, as shown in Figure~\ref{fig:mesh-architecture}(b). Alternative topologies, such as 3D torus or tree structures, are impractical due to high on-chip wiring costs. Therefore, wafer-scale chip makers such as Cerebras WSE~\cite{wse} and Tesla Dojo~\cite{dojo} adopt massive-scale mesh architectures. Even non-wafer-scale accelerators such as Meta MTIA~\cite{mtia}, Tenstorrent~\cite{tenstorrent}, and others~\cite{amd-xdna, azure-maia} use mesh to scale cores on a chip.

The massive-scale mesh architecture presents challenges for several LLM operations due to their high data movement demands: (i)~managing LLM models and KV cache, (ii)~GEMM operations during the prefill phase, and (iii)~GEMV operations during the decode phase. Other operations, such as element-wise computations such as dot-product and activation functions, require no data movement and naturally benefit from parallelism. Operations needing allreduce, such as RMSNorm and Softmax, can leverage GEMV solutions.

\vspace{-0.3cm}
\section{Device Model for Wafer-Scale Accelerators}
\vspace{-0.2cm}

\subsection{The PLMR model}
\vspace{-0.1cm}
We develop the PLMR model to capture the unique hardware properties of wafer-scale accelerators and to motivate system requirements needed for utilizing this emerging hardware.

\begin{enumerate}[label=(\arabic*), leftmargin=0.5cm, noitemsep,topsep=0pt]
    \item \textbf{Massive \underline{P}arallelism (P)}:
A wafer-scale accelerator can accommodate millions of parallel cores, compared to thousands in GPUs. Each core features a local hardware pipeline that overlaps data ingress, egress, computation, and memory access at the cycle level.
This requires the computation to be partitioned at a massive scale and a fine-grained schedule to overlap computation, memory access, and NoC communication.

\item \textbf{Highly non-uniform memory access \underline{L}atency (L)}: Accessing memory across cores in a mesh incurs highly non-uniform latency. In a mesh with $N_w \times N_h$ cores, organized as a rectangle, the maximum number of NoC hops between two cores is $N_w + N_h$, and the worst-case memory access latency is given by $\alpha(N_w + N_h) + \beta r$, where $r < N_w + N_h$ denotes the number of routing stages along the communication path. Here, $\alpha$ denotes the per-hop transmission latency, the cost incurred when a message is directly forwarded at a core according to the router hardware's pre-configured rules, increasing with hops. $\beta$ represents the per-routing latency, the overhead when a message is involved in header parsing and rewriting by software at a core when forwarding~\cite{wafer-reduce}. Typically, $\alpha < \beta$. In a mesh with a million cores, the maximum number of hops and routing can reach up to several thousand, resulting in up to a thousand times latency gap between local and remote memory access. Consequently, minimizing long-range communication is critical for performance.

\item \textbf{Constrained per-core local \underline{M}emory (M)}: Each core has a small local memory (tens of KBs to several MBs), as performance and energy efficiency decline with larger capacities~\cite{sram-wiki}. As a result, computation data must be explicitly partitioned into fine-grained chunks to fully fit within the constraints of each core's local memory.


\item \textbf{Constrained \underline{R}outing resources (R)}: Wafer-scale accelerators integrating millions of cores impose strict constraints on each core’s routing circuit complexity or routing table size. For example, on the Cerebras WSE-2, each core can recognize only message headers with a 5-bit address code. Consequently, each core can support at most $2^5$ distinct routing paths, and the software system must carefully plan these paths. For long-distance remote communication, a pair of cores can consume routing resources to establish a direct routing path, incurring only $\alpha$ latency. However, if the number of routing paths exceeds hardware limits, messages must be relayed through multiple intermediate cores, introducing additional $\beta$ latency.
\end{enumerate}

We expect these properties to remain relevant, as they are rooted in the fundamental characteristics of hardware and its manufacturing process. The PLMR model applies to both current (Cerebras WSE) and future (Tesla Dojo) wafer-scale devices. Even some non-wafer-scale devices with mesh-based NoC architectures, such as Tenstorrent Blackhole~\cite{tenstorrent}, can be represented by PLMR with adjusted parameters for parallelism (P), the size of the mesh (L), or relaxed constraints on local memory (M) and routing resources (R).

    \vspace{-3mm}
\subsection{Limitations of state-of-the-art approaches}
    \vspace{-1mm}
    
Leveraging the PLMR model, we analyze why existing AI systems fail to fully utilize wafer-scale accelerators.
To run an LLM on a wafer-scale accelerator, we generally have two choices: (i) abstract the distributed local memory in each core as a shared memory and directly access data placed in a remote core through NoC; and (ii) explicitly partition computation into distributed cores and use message passing to exchange necessary data. 
We analyze two types of representative systems: LLM runtime or DNN compilers for shared memory architecture such as GPUs, e.g., Ladder~\cite{ladder}; and the SOTA compiler for distributed on-chip memory architectures, e.g., T10~\cite{t10} for GraphCore IPU.

\mypar{Shared-memory system} 
A shared-memory-based DNN compiler such as Ladder usually assumes a uniform memory access pattern within the underlying memory hierarchy, which cannot tolerate the thousands of times latency variance in wafer-scale accelerators when accessing data from remote memory (failing in L).
Moreover, these compilers~\cite{tvm,rammer,ansor,flextensor,roller,welder,ladder} often focus primarily on partitioning computation, with less emphasis on optimizing data partitioning. This approach can easily lead to significant data duplication and violate the memory constraint requirements (failing in M).
Finally, these compilers are unaware of the communication distance of each core, poorly addressing the constraint of routing resources.

\mypar{Distributed-memory system} The T10 system~\cite{t10} is designed for AI accelerators with an on-chip crossbar, which ensures a constant latency of memory access to any other cores on the same chip. T10 handles small local memory and balances communication loads, addressing memory constraints (M) and routing resource limitations (R). However, on a PLMR device, it fails to account for varying memory access latency (failing in L) and scales to thousands, not millions, of cores (failing in P).

    \vspace{-3mm}
\section{Wafer-Scale LLM Parallelism}\label{sec:meshmp}
    \vspace{-1mm}
    
We present wafer-scale LLM parallelism, featuring new designs across prefill, decode, and KV cache management.

    \vspace{-3mm}
\subsection{Prefill parallelism}
    \vspace{-1mm}

The parallelism for LLM prefill must ensure compliance with the PLMR model. Key challenges include: (i)~Handling multiple large matrices during prefill, requiring effective dimension partitioning to achieve million-core parallelism (P); (ii)~Optimizing GEMM operations, which involve further partitioning and overlapping computation and communication, to minimize long-range communication latency (L), respect local memory constraints (M), and account for limited routing resources (R); and (iii)~Handling matrix transposes, which are costly on a mesh NoC (L) but often required for sequential GEMM operations.

\begin{figure}[t!]
    \centering
    \includegraphics[width=0.46\textwidth]{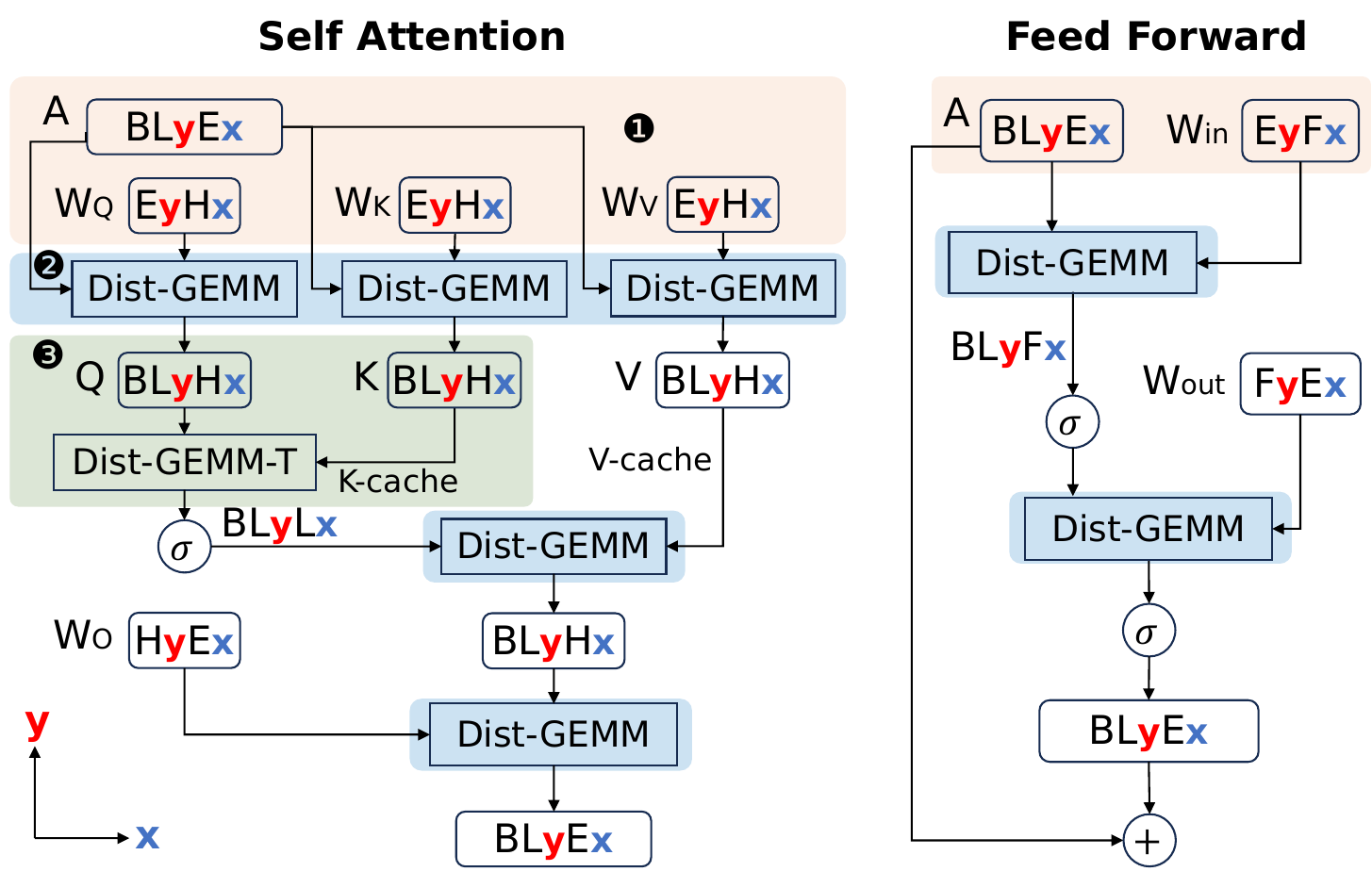}
    \vspace{-2mm}
    \caption{Prefill parallelism plan. $E_xF_y$ represents a matrix of shape $EF$, where the $E$ dimension is partitioned along the $x$-axis of cores, and $F$ along the $y$-axis of cores on a mesh.}
    \vspace{-5mm}
    \label{fig:prefill-partition}
\end{figure}

\mypar{Designing fine-grained partitioning for million-core parallelism} To achieve high chip utilization, we propose partitioning two dimensions of the input activation and weight matrices along both the $X$- and $Y$-axes of cores. This approach enables finer-grained, million-scale parallelism compared to existing methods~\cite{flashattn2, flashdecoding++, megatron, google2023efficiently}, which typically partition only the embedding dimension, resulting in insufficient parallelism on PLMR devices. 

We illustrate this partitioning using self-attention and feedforward, as shown in Figure~\ref{fig:prefill-partition}. For this discussion, we define the following annotations: the input activation $A$ and weight $W$ are multi-dimensional tensors during the prefill process. $B$ represents the batch size, $L$ the sequence dimension~\footnote{$L$: sequence dimension; L: memory access latency in PLMR.}, $E$ the embedding dimension, $H$ the head dimension, and $F$ the hidden dimension in the feedforward block.
As shown by \myc{1}, the partitioning layout of $A$ is represented as $BL_yE_x$, where the $L$ dimension is partitioned along the $Y$-axis of cores, and the $E$ dimension along the $X$-axis of cores. Similarly, all weight matrices ($W_Q$, $W_K$, $W_V$, $W_{in}$, and $W_{out}$) are partitioned across both dimensions.

\mypar{Designing PLMR-compliant distributed GEMM} 
We propose replacing conventional GEMM operators, which are designed for shared memory architectures, with a newly designed PLMR-compliant distributed GEMM during the prefill phase (as shown in \myc{2} of Figure~\ref{fig:prefill-partition}). Unlike TPU and GPU systems that primarily rely on allgather operations for GEMM, PLMR-compliant distributed GEMM algorithms achieve high NoC bandwidth utilization while respecting local memory and routing constraints, ensuring compliance with the L, M, and R properties. This PLMR-compliant distributed GEMM is fully described in Section~\ref{sec:gemm}.

\mypar{Using transposed distributed GEMM to avoid matrix transpose}
We propose a transpose-free parallelism plan for prefill to avoid matrix transpose, a common operation in LLM systems designed for shared memory architectures. The L property in PLMR highlights that matrix transposition is particularly costly on a wafer-scale device. It requires a core on one corner of the mesh to send data to the opposite diagonal corner, creating a long-range communication path.

Our transpose-free parallelism plan leverages transposed distributed GEMM (denoted as dist-GEMM-T)~\cite{summa, trans-dist} to compute $Q@K^T$ during LLM prefill, as shown by \myc{3} in Figure~\ref{fig:prefill-partition}. Specifically, the intermediate $Q$ and $K$ tensors, generated by multiplying $X$ with $W_Q$ and $W_K$, require transposing $K$ before proceeding with dist-GEMM operations due to the on-chip partition shape.

    \vspace{-3mm}
\subsection{Decode parallelism}
    \vspace{-1mm}

\begin{figure}[t!]
    \centering
    \includegraphics[width=0.46\textwidth]{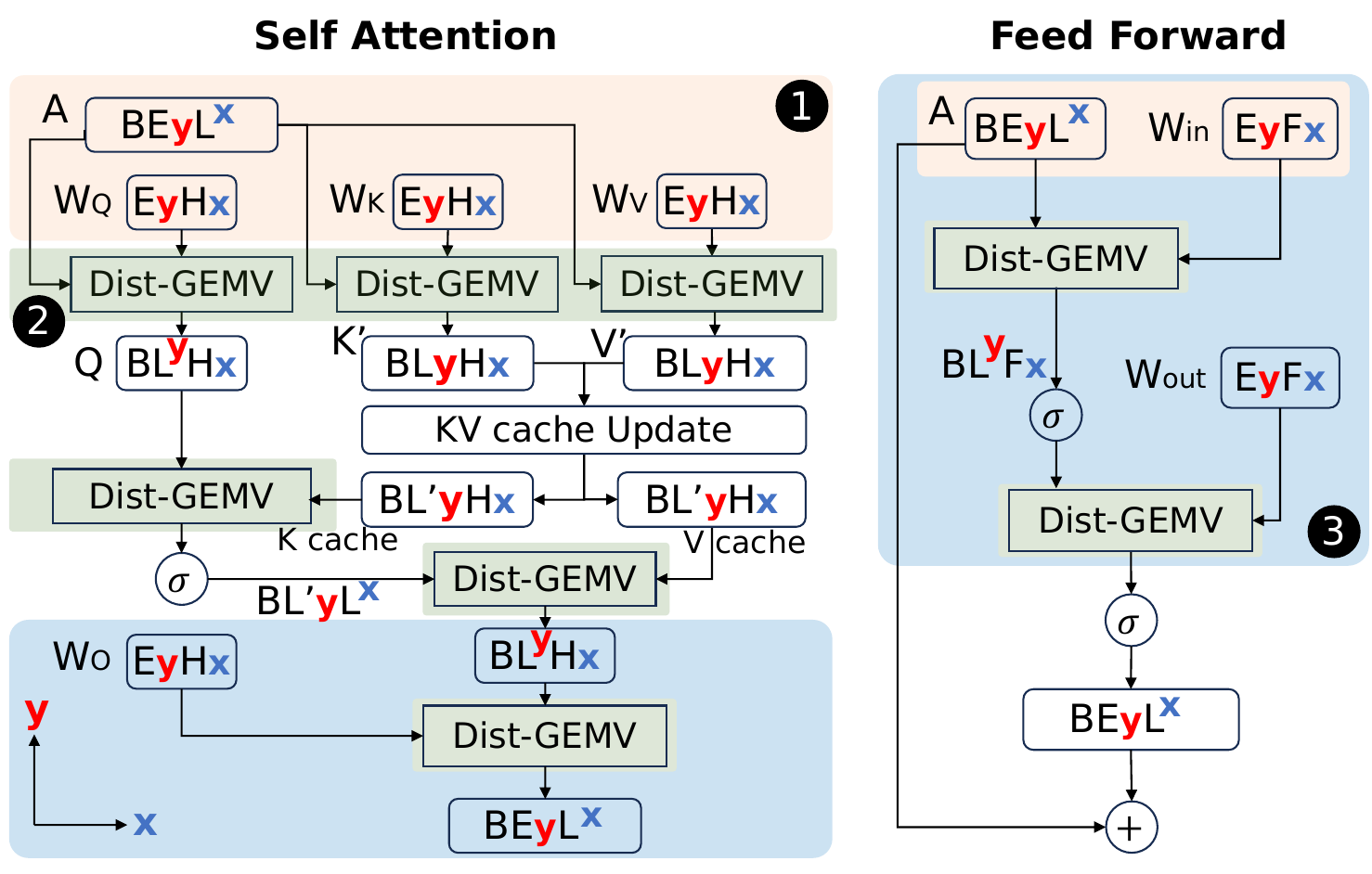}
    \vspace{-2mm}
    \caption{Decode parallelism plan. $E^yF_x$ indicates the $E$ dimension is replicated along the $y$-axis, and $F$ is partitioned along the $x$-axis.}
    \vspace{-5mm}
    \label{fig:decode-partition}
\end{figure}

The parallelism strategy for LLM decode must address its memory-bandwidth-intensive nature, presenting several challenges: (i)~Decode uses smaller matrices than prefill due to limited input sequences and batch sizes, requiring careful parallelization when dimensions are insufficient for partitioning; (ii)~The phase heavily relies on GEMV operations, which are less compute-intensive than GEMM, resulting in short computation phases with limited overlap with communication, making GEMV vulnerable to long-range communication latency on a mesh NoC (L) and requiring adherence to local memory and routing constraints (M and R); and (iii)~Sequential GEMV operations introduce costly matrix transpose on a NoC, risking violation of the L property.

\mypar{Designing fine-grained replication to enable parallelism at minimal communication cost} When tensor dimensions are insufficient to achieve the high parallelism required for decode, we propose fine-grained replication of tensors in LLMs, specifically replicating the sequence dimension, where the sequence length equals the prompt length during prefill phase and equals 1 during the decode phase. This approach offers two key advantages: (i)~ it improves parallelism and ensures balanced loads across all cores, and (ii)~ it avoids additional communication operations such as allreduce across all cores. As shown by \circled{1} in Figure~\ref{fig:decode-partition}, the $E$ dimension is partitioned along the $y$-axis, and the $L$ dimension is replicated along the $x$-axis, represented as $BE_yL^x$. Weight matrices $W$ are partitioned across both dimensions, consistent with the prefill phase.

Our fine-grained replication differs from recent work on long-context/sequence inference systems~\cite{loongserve, distserve}, which selectively replicate certain dimensions during the prefill phase rather than the decode phase.

\mypar{Designing PLMR-compliant distributed GEMV} We found that existing GEMV implementations fail to fully comply with PLMR requirements due to long-range communication and excessive routing resource consumption at each core. To address this, we propose a PLMR-compliant distributed GEMV, utilizing this new implementation throughout the decode phase (as detailed in \myc{2} of Figure~\ref{fig:decode-partition}). A comprehensive description of this GEMV design is provided in Section~\ref{sec:gemv}.

\mypar{Pre-optimizing model weight placement to avoid matrix transpose}
To avoid matrix transpose during decode, we pre-optimize the model weight layout for decode, particularly for the distributed GEMV operation, to eliminate matrix transpose. While this introduces re-placement overhead between prefill and decode phases, the overhead is far smaller than that of sequential matrix transpose during token generation.

Figure~\ref{fig:decode-partition} illustrates this proposal, detailed in \myc{3}. Specifically, we optimize the placement of weights such as $W_{O}$ and $W_{out}$ for distributed GEMV in decode, differing from their layout in the prefill phase. This approach also removes transpose operations in calculating $Q@K^T$ during decode self-attention.

    \vspace{-3mm}
\subsection{Shift-based KV cache management}\label{sec:parallel:kvcache}
    \vspace{-1mm}

\begin{figure}[t!]
    \centering
    \includegraphics[width=1\linewidth]{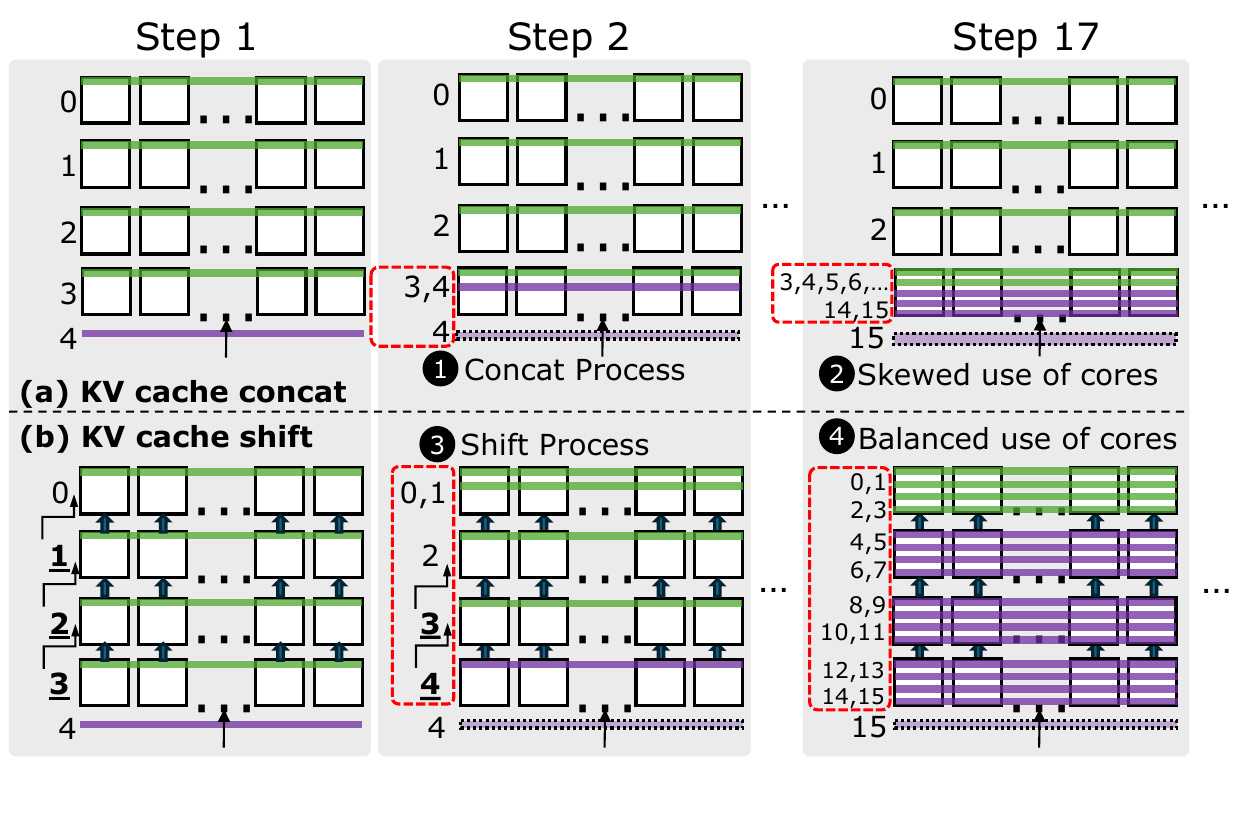}
    \vspace{-9mm}
    \caption{KV cache concatenation vs. KV cache shift}
    \vspace{-5mm}
    \label{fig:kv_cache_update}
\end{figure}

KV cache management on PLMR devices is challenging as it requires storing large data across distributed cores while adhering to local memory constraints (M) and distributing KV cache computations to achieve high parallelism (P). To address these, we have the following insights:

\mypar{Existing concatenate-based management causes skewed core utilization} Current KV cache management methods primarily concatenate the most recently generated KV vectors to the existing cache. Though efficient in shared memory architectures, this concatenate operation leads to highly skewed core utilization on PLMR devices, as shown in \circled{1} of Figure~\ref{fig:kv_cache_update}, where only the core in a row is responsible for storing and computing over the generated KV vector. After several token generation steps, this only core quickly becomes the bottleneck, as depicted in \circled{2} of Figure~\ref{fig:kv_cache_update}, causing skewed memory usage and violating the M in PLMR. Moreover, the imbalanced KV cache distribution across cores results in inefficient parallelism, violating the P property.

\mypar{Proposing shift-based management for balanced core utilization} We propose a shift-based KV cache management strategy that evenly distributes cache data across all cores. Instead of concatenating new KV cache vectors at the end, this method performs a balancing shift operation, where each row transfers the oldest KV cache data to the row above, as shown in \circled{3} of Figure~\ref{fig:kv_cache_update}. When new KV data arrives, each core checks its local capacity against its neighbors. If equal, upward shifts are triggered, with each row receiving data from below and passing some to the row above. As illustrated in \circled{4}, this ensures even KV cache distribution across all cores.

The upward shifts utilize NoC links in parallel, maintaining high performance and satisfying the P property. The physical placement of KV cache aligns with logical continuity and only involves data movement between the adjacent cores, adhering to the L property. This method also fully resolves the M violation issue observed in the last row of cores with the concatenate-based approach.

    \vspace{-3mm}
\subsection{Implementation details}
    \vspace{-1mm}

We outline several implementation details below:

\mypar{Prefill and decode transition} Prefill and decode require distinct strategies. To handle the transition efficiently, we reshuffle KV cache and weights through the fast NoC, which often provides 100s Pbits/s aggregated bandwidth, completing instantly without relying on slower off-chip memory.

\mypar{Parallelism configuration} \sys uses offline autotuning to select core counts for each model, optimizing latency based on model size, input/output length, memory per core, and prefill/decode phases. It chooses different core counts for each phase, with fast dynamic remapping enabled by high NoC bandwidth. For models with variable input/output lengths, average values are used to maintain near-peak performance. Autotuning runs separately per model to adapt to specific needs.

\mypar{Variations of self-attention} \sys supports variations of Self-Attention, including Grouped Query Attention~\cite{gqa}, Multihead Attention~\cite{mha}, and Multi-query Attention~\cite{mqa}. These differ by performing dist-GEMM, dist-GEMV and dist-GEMM-T locally after grouping by head dimensions.

    \vspace{-3mm}
\section{Wafer-Scale GEMM} \label{sec:gemm}
    \vspace{-1mm}

In this section, we introduce \gemm, a scalable distributed GEMM for massive-scale, mesh architectures. 

    \vspace{-3mm}
\subsection{PLMR compliance in distributed GEMM}
    \vspace{-1mm}

\begin{figure}
    \centering
    \includegraphics[width=1\linewidth]{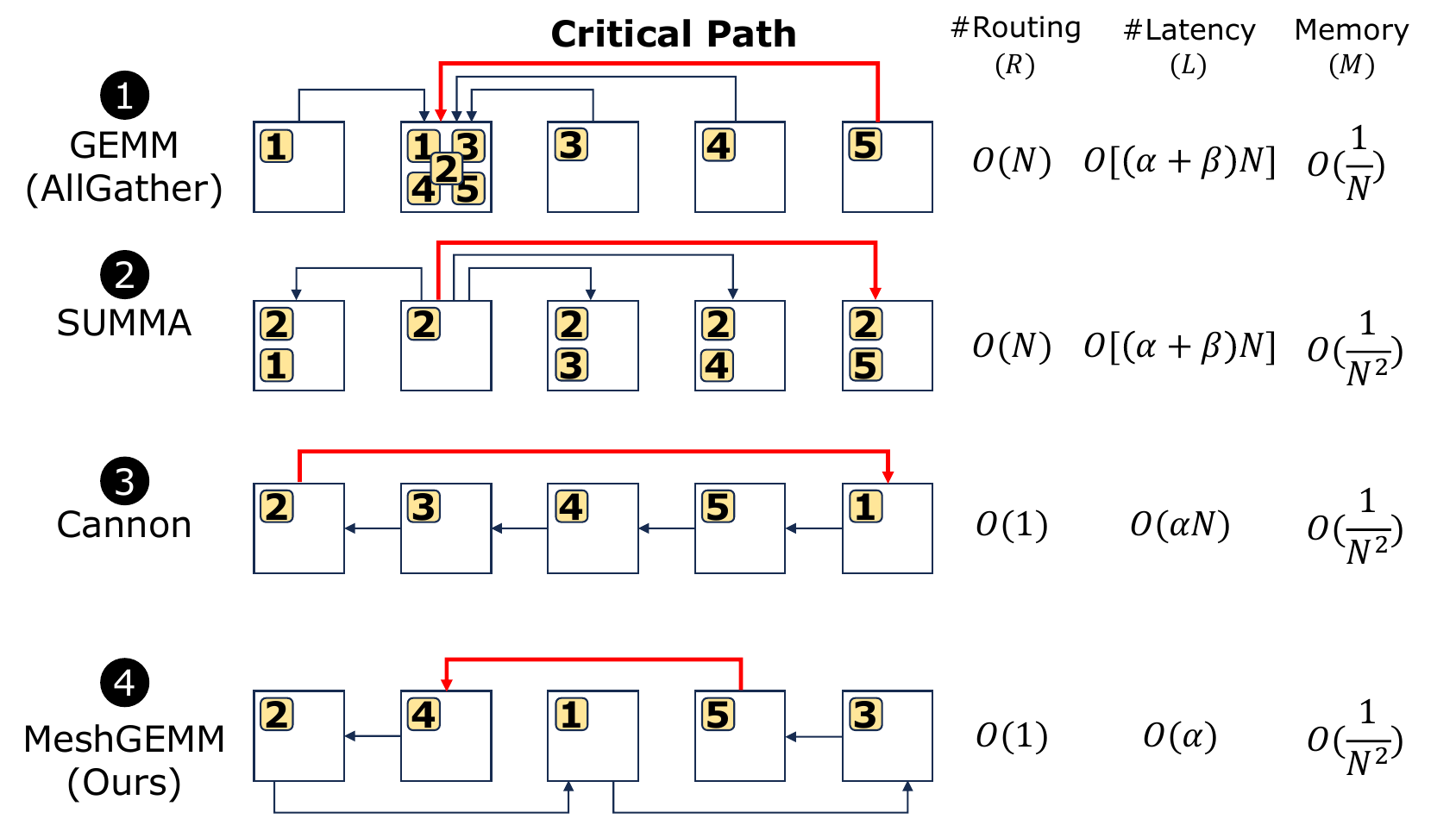}
    \vspace{-3mm}
    \caption{PLMR compliance in distributed GEMM}
    \vspace{-3mm}
    \label{fig:distributed-gemm-analysis}
\end{figure}

To identify a scalable distributed GEMM for PLMR devices, we define the following metrics: 
(i)~\emph{Routing paths per core}: The number of routing paths per core, with fewer paths ensuring compliance with the R property.
(ii)~\emph{Latency of critical path}: Maximal latency among all communication paths in each step to transmit submatrix (as the red lines in Figure~\ref{fig:distributed-gemm-analysis}), with less latency adhering to the L property.
(iii)~\emph{Memory per core}: The memory required per core, with lower usage ensuring the M property.

We analyze current distributed GEMM methods and show how \gemm meets these metrics:

\begin{enumerate}[label=(\arabic*), leftmargin=0.5cm, noitemsep,topsep=0pt]
\item \textbf{GEMM via Allgather} is commonly used in GPU and TPU pods for distributed GEMM~\cite{google2023efficiently, tensorrt-llm, megatron}. Its critical communication path in each step is one core gathering data from the farthest cores, shown as the red line in Figure~\ref{fig:distributed-gemm-analysis}~\circled{1}, and $N$ steps to complete the allgather. Each core creates $N$ routing paths to neighbors in its row and column (violating R). The constraint of $R$ necessitates step-by-step submatrix transmission via intermediate cores, introducing a communication latency of $O[(\alpha + \beta)N]$ along the critical path (violating L), and each core uses $O(1/N)$ memory due to inflated working buffers (violating M).

\item \textbf{SUMMA} is Cerebras' default choice for distributed GEMM on its wafer-scale engine~\cite{cerebrasgemm}. Its critical communication path in each step is where one core broadcasts data to the farthest core along the column or row, shown by the red line in \circled{2} of Figure~\ref{fig:distributed-gemm-analysis}. Same to the allgather communication in~\circled{1}, each core creates $N$ routing paths (violating R) and spans the critical path with $O[(\alpha + \beta)N]$ latency (violating L) in each step. While SUMMA improves memory usage compared to allgather, requiring only a working set equal to the size of locally partitioned submatrices, it still doubles the peak memory usage.

\item \textbf{Cannon} is mesh-optimized choice for distributed GEMM~\cite{cannon}, popular in supercomputers and distributed cluster. Its critical communication path in each step is the head cores send data to the tail cores shown in~\circled{3} of Figure~\ref{fig:distributed-gemm-analysis}. Each intermediate core communicates with two neighbours in a 2D torus and passes through the submatrix from head to tail, which only needs $O(1)$ communication paths and optimal memory usage of $O(1/N^2)$. 

Notably, Cannon involves only a constant number of routing paths per core, allowing static routing rules to be assigned for both neighbor communication and the critical path. At each step, submatrices can pass directly from the head to the tail core through intermediate cores, unlike GEMM via allgather or SUMMA, which require step-by-step message transmission. As a result, the critical path incurs only per-hop latency $\alpha$, without additional per-routing overhead $\beta$. However, since the critical path spans $N$ hops, it still incurs $O(\alpha N)$ latency, shown as the red line in \circled{3}, violating the $L$.

\item \textbf{MeshGEMM (Ours)} is a distributed GEMM which complies with the PLMR model. The critical communication path in each step, which we named as two-hop transmission, is shown as the red line in \circled{4} of Figure~\ref{fig:distributed-gemm-analysis}, which is significantly shorter than the others. Each core communicates with two two-hop away neighbors(proven in later sections to be scalable for larger-sized mesh architectures), and passes through a one-hop neighbor's communication. This design achieves $O(1)$ communication paths per core needed and optimal memory usage of $O(1/N^2)$, similar to Cannon. Crucially, it bounds the critical path to constantly 2 hops with $O(\alpha)$ complexity, making it uniquely capable of addressing the L property.

\end{enumerate}

\vspace{-0.2cm}
\subsection{Design intuitions and scalability analysis} 

Our design involves two phases: (i)~We ensure algorithm correctness using a cyclic shifting process for GEMM, and (ii)~ then use interleaving to bound the critical path latency to a constant.
Based on the cyclic shifting and interleaving, we get the \textbf{two-hop transmission} communication path, and we prove that it, on this cycle, is the minimal distance required to satisfy the L property.

\begin{figure}[t!]
    \centering
    \includegraphics[width=1\linewidth]{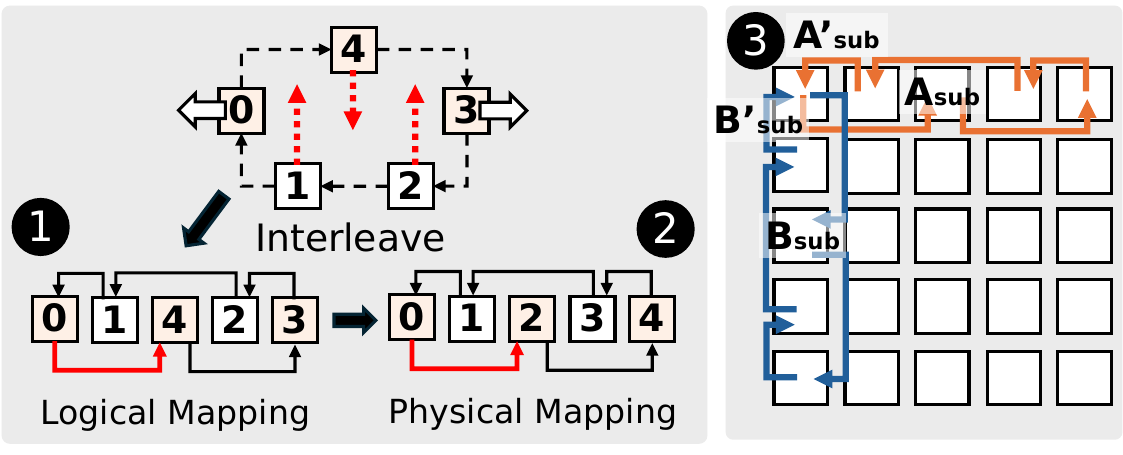}
    \caption{Design intuitions and scalability analysis.}
    \label{fig:gemm_intuition}
\end{figure}

\mypar{Cyclic shifting} Cyclic shifting enables \gemm to satisfy the M and R properties by limiting communication to two neighbors and minimizing memory usage. It ensures correct GEMM results, following reasoning similar to Cannon~\cite{cannon}. As illustrated in \circled{3} of Figure~\ref{fig:distributed-gemm-analysis}, a logical circle of 5 cores is flattened into the physical communication mapping, with a critical path from head core to tail core. 

\begin{algorithm}[t]
    \caption{$\textbf{INTERLEAVE}$}
    \label{alg:interleave}
    \SetKwFunction{assert}{assert}
    \SetKw{ret}{Return}
    \SetKw{int}{Int}
    \LinesNumbered
    \KwIn{index, N}
    \KwOut{send\_index, recv\_index}
    
    \eIf{\textnormal{index} \textnormal{\textbf{mod}} \textnormal{2 == 0}}{
        recv\_index = \texttt{Max} (index - 2, 0)\;
        send\_index = \texttt{Min} (index + 2, N - 1)\;
    }{
        recv\_index = \texttt{Min} (index + 2, N - 1)\;
        send\_index = \texttt{Max} (index - 2, 0)\;
    }
    \textbf{if} \textnormal{index == 0} \textbf{then} recv\_index = 1\;
    \If{\textnormal{index == N - 1}}{
        \textbf{if} N \textbf{mod} 2 == 0 \textbf{then}
            recv\_index = N - 2\;
        \textbf{else}
            send\_index = N - 2\;
    }
    \ret send\_index, recv\_index\;
\end{algorithm}

\mypar{Interleaving} For the flattened communication plan, we would like to minimize the length of the critical path further, thus satisfying the L property. Our key intuition here is to introduce an INTERLEAVE operation to find the mapping relationship from logical to physical, defined in Algorithm~\ref{alg:interleave}. As shown by \circled{1} of Figure~\ref{fig:gemm_intuition}, \gemm first insert core 1 in between core 0 and 4 and core 2 in between core 4 and 3 to form a logical mapping, and then call the INTERLEAVE operation to get the send to and receive from neighbours' index, resulting in a permutated, equivalent communication plan as shown by \circled{2} in Figure~\ref{fig:gemm_intuition}. For example, there are 5 cores total (N=5), so physical core 2 (index=2) sends data to physical core 4 (send\_index=4) and receives from physical core 0 (recv\_index=0). Figures \ref{fig:distributed-gemm-analysis} and \ref{fig:gemm_intuition} illustrate the case with 5 cores as an example, while Algorithm \ref{alg:interleave} demonstrates that the two-hop communication pattern in \gemm generalizes to mesh architectures of arbitrary size with $N \geq 3$.

\mypar{Scalability analysis}  We can prove that the two-hop distance created by INTERLEAVE cannot be further reduced. The proof relies on the fundamental properties of sequential arrangements: if we attempt to create a circular sequence where each number differs from its neighbors by exactly one hop, we encounter a mathematical impossibility. This can be understood by visualizing the numbers as points on a line - while adjacent numbers can be connected, the endpoints of the sequence cannot simultaneously maintain single-hop differences with their neighbors while forming a circle.

Note that our discussion, based on a 1D array, naturally extends to a 2D mesh, as the 1D array corresponds to the mesh's X-axis and Y-axis due to their symmetry.

\vspace{-0.3cm}
\subsection{The \gemm algorithm} \label{subsec:meshgemm} 

We outline the key steps of \gemm below:

\begin{enumerate}[label=(\arabic*), leftmargin=0.5cm, noitemsep,topsep=0pt]
\item \textbf{Initialization:} Consider $C = A \times B$. \gemm will partition $A$ and $B$ into tiles $A_{sub}$ and $B_{sub}$ along two dimensions, forming $N \times N$ tiles, which are distributed across the cores. Each core receives one tile of $A_{sub}$ and one of $B_{sub}$. \gemm will then use INTERLEAVE to initialize the neighbors' positions for each core. 

\item \textbf{Alignment:}  
Each core will then align with neighbors to align the input submatrices in a way that ensures every core in the distributed system begins with the appropriate operands for the matrix multiplication process. 

\item \textbf{Compute-shift loop:}  
Each core operates with a compute-shift loop involving $N$ steps of communication and computation. In each step, every core computes the partial sum of its corresponding $C_{sub} = A_{sub} \times B_{sub} + C_{sub}$. Meanwhile, shift $A_{sub}$ along the X-axis and $B_{sub}$ along the Y-axis to get new $A'_{sub}$ 
 and $B'_{sub}$ for the next step computation as \circled{3} we shown in Figure~\ref{fig:gemm_intuition}. After $N$ steps, the accumulated $C_{sub}$ is returned.
\end{enumerate}

\vspace{-0.3cm}
\subsection{Implementation details}

\mypar{Handling non-square mesh} For a non-square mesh $N_h \times N_w$ ($N_h \neq N_w$), the $A$ and $B$ matrices can be logically partitioned into $N_{lcm} \times N_{lcm}$ cores, where $N_{lcm}$ is the least common multiple of $N_h$ and $N_w$.

\mypar{Transposed distributed GEMM} The above algorithm 
can be applied to the computation of $C = A \times B^T$, the dist-GEMM-T in Figure~\ref{fig:prefill-partition} to avoid transposing $B$ on mesh. It does not require alignment before computation and only necessitates $N$ steps of two-hop compute-shift for the right matrix $B$ along the Y-axis. After each shift step, each core computes $C_{\text{sub}} = A_{\text{sub}} \times B_{\text{sub}}$, followed by a ReduceAdd of $C_{\text{sub}}$ along the X-axis. After $N$ steps, the final matrix $C$ is obtained.

\vspace{-0.3cm}
\section{Wafer-Scale GEMV}\label{sec:gemv}
\vspace{-0.2cm}

In this section, we describe \gemv, a scalable GEMV algorithm for PLMR devices.

\subsection{PLMR compliance in distributed GEMV}

\begin{figure}[t!]
    \centering
    \includegraphics[width=1\linewidth]{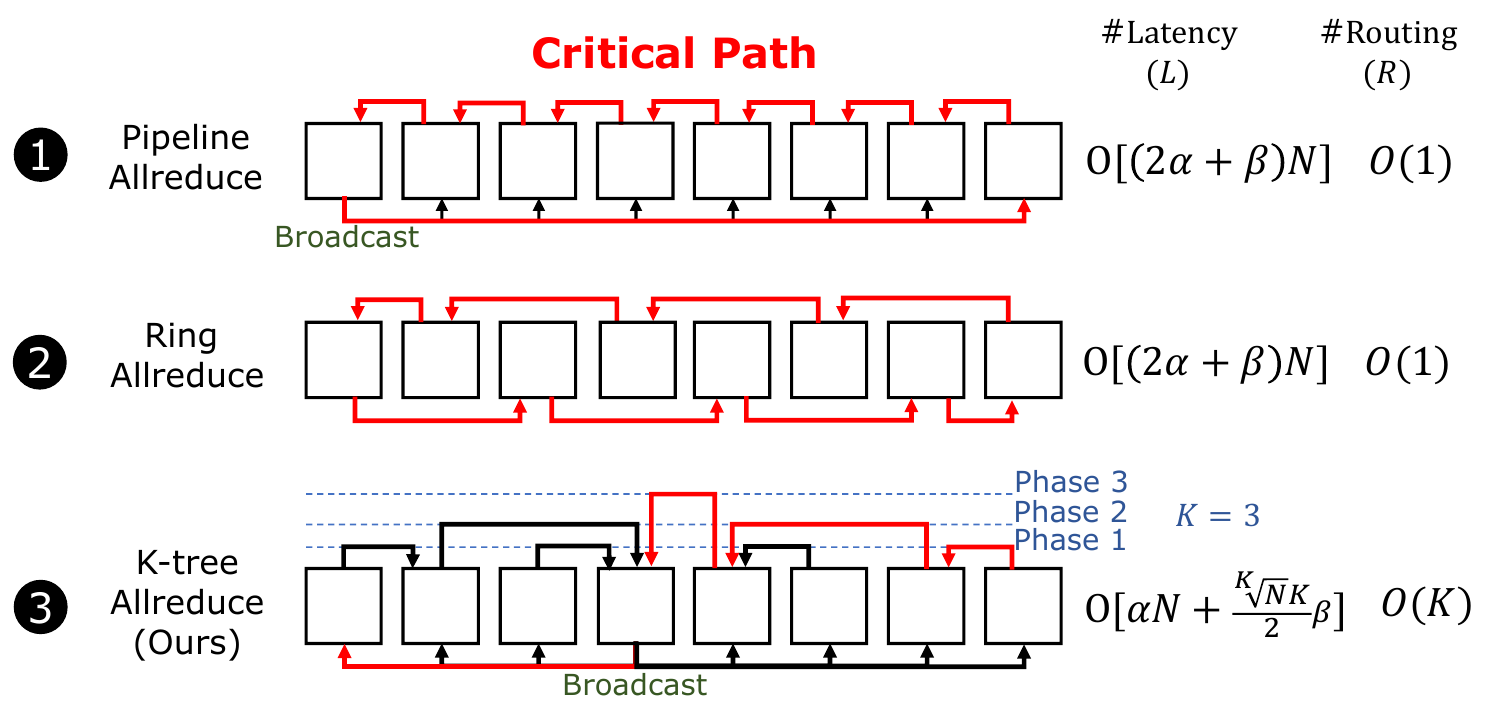}
    \caption{PLMR compliance in distributed GEMV}
    \label{fig:gemv-plmr-compliance}
\end{figure}

The completion time of a distributed GEMV is primarily determined by an element-wise computation to generate a partial sum result at each core and then an allreduce operation that aggregates partial results from all selected cores and then sends the aggregated results back to all cores for the continuous GEMV. As the analysis in GEMM, in wafer-scale GEMV, we also define the following metrics: 
(i)~\emph{Routing paths per core}: The number of routing paths per core, with fewer paths ensuring compliance with the R property
(ii)~\emph{Latency of critical path}: Maximal latency of the whole allreduce communication, mainly determined by the time from when the partial sum is sent from the farthest core to when the aggregated sum is received by the farthest core.

We analyze current distributed GEMV methods and show how \gemv meets these metrics:

\begin{enumerate}[label=(\arabic*), leftmargin=0.5cm, noitemsep,topsep=0pt]

 \item \textbf{GEMV with pipeline allreduce} is commonly used in TPU pod systems~\cite{google2023efficiently} and as the default in Cerebras demo~\cite{cerebrasgemv}. As illustrated by the red line in the Figure \ref{fig:gemv-plmr-compliance}~\circled{1}, the critical path of the pipelined all-reduce starts from the farthest core, where partial sums are reduced step-by-step toward the root core, and the aggregated result is then broadcast from root to all cores. While this approach ensures that the routing path per core remains within the $R$ constraint, $O(1)$ per-core routing resource usage, it incurs a total of $2N$ hops and $N$ routing stages, violating the L property.

\item \textbf{GEMV with ring allreduce} is commonly used in GPU pod systems, where it is the default configuration, especially for a large amount of data \cite{wafer-reduce}. As shown by \circled{2} in Figure~\ref{fig:gemv-plmr-compliance}, the critical path of ring allreduce involves each partial sum traversing all cores in the ring. Similar to pipelined allreduce, this approach maintains $O(1)$ routing paths per core, but incurs a communication latency of $O\left[(2\alpha + \beta)N\right]$, thus violating the latency constraint $L$ defined in PLMR.

\item \textbf{GEMV with K-tree allreduce (Ours)}. As analyzed above for pipelined and ring allreduce, the sequential execution of reduce-add operations leads to $N$ routing stages along the critical path. In contrast, K-tree allreduce organizes the reduce-add path as a balanced K-tree, enabling $K$ phases grouped parallel reductions with $O(\sqrt[K]{N})$ cores per group and reducing the critical path to only $\frac{\sqrt[K]{N}K}{2}$ times routing and $N$ hops. However, this comes at the cost of requiring $O(K)$ routing paths per core.

\end{enumerate}

\vspace{-3mm}
\subsection{The \gemv algorithm}
\vspace{-1mm}

We will outline the key steps of \gemm below:

\begin{enumerate}[label=(\arabic*), leftmargin=0.5cm, noitemsep,topsep=0pt]
\item \textbf{Initialization:} Consider $C = A \times B$ and $A$ is a vector. \gemv will partition $B$ into tiles $B_{sub}$ along two dimensions, forming $N \times N$ tiles and distributed across the cores. For $A$, \gemv will partition it along the vector length, forming $N$ tiles distributed on one axis and replica $A$ on another axis. Each core receives one tile of $A_{sub}$ and one of $B_{sub}$. Then we determine which cores form a group to obtain aggregated results in each phase based on the K-tree.

\item \textbf{Parallel computation:} In this stage, each core performs a local GEMV $A_{\text{sub}} \times B_{\text{sub}}$ to obtain $C_{sub}$ partial sum.

\item \textbf{Aggregation:} The aggregation step primarily involves using the K-tree allreduce we design. The key steps are as follows: (i)~In the 1st-phase, each group performs group reduction and obtains the partial sum of $C_{sub}$ at the root core of each group. (ii)~In the $k$th-phase, the results from the $(k-1)$~th-phase are reduced to the root cores of each group in the $k$th-phase. After K times repeating, $C$ can be obtained by concatenating the $C_{sub}$ from all K-tree root cores. (iii)~Optionally, a broadcast operation from the root core of the $K$-tree may follow, depending on whether continuous GEMV is required.
\end{enumerate}

\mypar{Scalability Analysis}  
As shown in \circled{3} of Figure~\ref{fig:gemv-plmr-compliance}, this method scales efficiently with parallelism and meets the L property by selecting an appropriate $K$. It requires $K+1$ paths at the tree root core but allows flexible adjustment of $K$ to address R based on hardware limitations. Compared to pipeline and ring all-reduce, K-tree all-reduce essentially builds pass-through paths between distant cores by consuming routing resources, thereby reducing routing latency.

However, a larger $K$ is not always better, as it depends on $N$ and R constraints. Additionally, larger $K$ increases routing complexity and overhead. Considering these factors, we have chosen $K=2$ for our current implementation, evaluated in the following sections.

\vspace{-0.3cm}
\section{Evaluation}
\vspace{-0.2cm}

We extensively evaluated \sys against various state-of-the-art methods and systems. Our results show that:
\begin{enumerate}[label=(\arabic*), leftmargin=0.5cm, noitemsep,topsep=0pt]
\item \sys achieves orders of magnitude speedup over T10 and Ladder in LLM inference (\S\ref{sec:eval:e2e});
\item \sys{}'s 
\gemm and \gemv perform and scale strongly over the state-of-the-art (\S\ref{sec:eval:meshgemm});
\item \sys's shift-based KV cache management enables over 360-385$\times$ more token capacity (\S\ref{sec:eval:kvcache});
\item \sys on Cerebras WSE-2 achieves 10-20$\times$ e2e speedup compared to the optimal performance of SGLang on A100 GPUs-cluster, and about 30-40$\times$ speedup compared to a single A100 GPU. \sys also achieves approximately 2.5$\times$ better energy efficiency (\S\ref{sec:eval:e2e},\S\ref{sec:eval:gpu}).
\end{enumerate}

\mypar{Experiment setup}
We evaluate \sys on a server with Cerebras WSE-2. WSE-2 has 850,000 Cores, each with a Compute Engine (CE) operating at a maximum of 1.1 GHz. Each clock cycle can fetch two 32-bit operands from SRAM, perform a multiply-accumulate operation, and then write back to SRAM. Each core also has a fabric router that can send or receive 32-bit messages from neighbouring cores with a single clock cycle. Additionally, each core contains 48KB of SRAM, with the chip totalling 40GB of aggregated SRAM~\cite{wse}.

We compare \sys with two DNN compilers: (i)T10\cite{t10}, the state-of-the-art compiler for AI accelerators with inter-core connections and distributed on-chip memory, and (ii)Ladder\cite{ladder}, the state-of-the-art compiler for shared memory architectures. For T10, we implemented it on WSE-2, treating each core as part of a distributed memory system interconnected by a crossbar, despite the actual mesh topology. T10 maps data to core IDs and fetches data from local SRAM as required. For Ladder, we treated the distributed memory architecture of the chip, interconnected by mesh, as unified memory, requiring collective communication over the NoC to access data.

We use the Nvidia A100 for GPU comparison experiments, which shares the same 7nm process as the WSE-2 for fairness. The experiments utilize up to 16 GPUs across two nodes (2$\times$8 configuration). Within a single node, the eight A100 GPUs are interconnected via NVLink, while inter-node communication is handled through a high-performance InfiniBand (IB) network. We use SGLang\cite{sglang}, one of the highest-performing LLM inference systems for GPU-based experiments.

\mypar{Experiment metric} To evaluate the maximum per-request throughput performance of LLM inference systems on different hardware, we define Throughput per Request (TPR) as a key metric. TPR is derived from the more widely used Time per Output Token (TPOT), with $\text{TPR} = \frac{1}{\text{TPOT}}$.

\mypar{LLM models}
Our evaluation includes various representative LLMs of different sizes and architectures. Specifically, LLaMA3-8B and LLaMA2-13B are widely used open-source LLMs, with LLaMA3 using group-query attention instead of multi-head attention to reduce KV cache usage. For LLaMA2-13B, we modified the model to remove the 4K context length limitation to evaluate system performance across different input and output lengths. Additionally, due to the tensor parallelism constraints, the number of attention heads of a model must be divisible by the number of GPUs used. We did not conduct the multi-GPU experiment of LLaMA-2 on a 16-GPU cluster. CodeLLaMA-34B is a specialized LLM for coding tasks, while QWen2-72B, another popular LLM, is renowned for its high model quality.

\vspace{-0.2cm}
\subsection{LLM inference} \label{sec:eval:e2e}

We first report the end-to-end performance of \sys compared to T10 and Ladder on WSE-2 and SGLang on the A100 GPU cluster. Then we further analyze the performance to provide deeper insights by breaking down the execution into prefill and decode phases.

\begin{table}[t]
  \centering
  \resizebox{\linewidth}{!}{
    \begin{tabular}{l|cc|cccc}
\hline
\multicolumn{1}{c|}{\multirow{2}{*}{Model}} & \multicolumn{2}{c|}{\multirow{2}{*}{Device}} & \multicolumn{4}{c}{Input/Output Seqence Length} \\ \cline{4-7} 
\multicolumn{1}{c|}{} & \multicolumn{2}{c|}{} & \multicolumn{1}{c|}{2048/128} & \multicolumn{1}{c|}{4096/128} & \multicolumn{1}{c|}{2048/2048} & 4096/4096 \\ \hline
\multirow{6}{*}{LLaMA3-8B} & \multicolumn{1}{c|}{\multirow{3}{*}{WSE-2}} & \sys & \multicolumn{1}{c|}{764.4} & \multicolumn{1}{c|}{604.4} & \multicolumn{1}{c|}{2370.3} & 2459.0 \\
 & \multicolumn{1}{c|}{} & T10 & \multicolumn{1}{c|}{4.6} & \multicolumn{1}{c|}{4.5} & \multicolumn{1}{c|}{58.3} & 94.6 \\
 & \multicolumn{1}{c|}{} & Ladder & \multicolumn{1}{c|}{1.2} & \multicolumn{1}{c|}{1.1} & \multicolumn{1}{c|}{7.4} & 8.7 \\ \cline{2-7} 
 & \multicolumn{1}{c|}{\multirow{3}{*}{\begin{tabular}[c]{@{}c@{}}A100 GPU\\ (SGLang)\end{tabular}}} & 1 & \multicolumn{1}{c|}{34.8} & \multicolumn{1}{c|}{31.1} & \multicolumn{1}{c|}{36.5} & 78.4 \\
 & \multicolumn{1}{c|}{} & 8 & \multicolumn{1}{c|}{117.2} & \multicolumn{1}{c|}{109.0} & \multicolumn{1}{c|}{128.4} & 256.1 \\
 & \multicolumn{1}{c|}{} & 2$\times$8 & \multicolumn{1}{c|}{73.7} & \multicolumn{1}{c|}{70.2} & \multicolumn{1}{c|}{79.3} & 162.5 \\ \hline
\multirow{5}{*}{LLaMA2-13B} & \multicolumn{1}{c|}{\multirow{3}{*}{WSE-2}} & \sys & \multicolumn{1}{c|}{473.9} & \multicolumn{1}{c|}{414} & \multicolumn{1}{c|}{1690.3} & 1826.0 \\
 & \multicolumn{1}{c|}{} & T10 & \multicolumn{1}{c|}{2.6} & \multicolumn{1}{c|}{2.5} & \multicolumn{1}{c|}{35.0} & 58.3 \\
 & \multicolumn{1}{c|}{} & Ladder & \multicolumn{1}{c|}{0.7} & \multicolumn{1}{c|}{0.7} & \multicolumn{1}{c|}{4.9} & 6.1 \\ \cline{2-7} 
 & \multicolumn{1}{c|}{\multirow{2}{*}{\begin{tabular}[c]{@{}c@{}}A100 GPU\\ (SGLang)\tablefootnote{\label{missing_16GPU}No 2$\times$8 GPUs, due to model architecture and tensor parallelism.}\end{tabular}}} & 1 & \multicolumn{1}{c|}{20.4} & \multicolumn{1}{c|}{17.1} & \multicolumn{1}{c|}{21.1} & 47.9 \\
 & \multicolumn{1}{c|}{} & 8 & \multicolumn{1}{c|}{79.6} & \multicolumn{1}{c|}{70.5} & \multicolumn{1}{c|}{86.9} & 172.4 \\ \hline
\end{tabular}%
    }
  \caption{End-to-end LLM inference TPR}
  \label{tab:e2e_table}%
\end{table}%

\mypar{End-to-end throughput} Table~\ref{tab:e2e_table} shows the inference TPR of LLaMA3-8B and LLaMA2-13B on WSE-2 and A100 with different input and output sequence lengths. Here, the end-to-end throughput is calculated as the total number of tokens generated during the decode phase divided by the total time spent in the prefill and decode phases. \sys uses core configurations optimized for the best performance with each model.
In LLaMA3-8B, we use 660$\times$660 cores for prefill and 360$\times$360 for decode.
In LLaMA2-13B, we use 750$\times$750 cores for prefill and 375$\times$375 for decode.
CodeLLaMA-34B and QWen2-72B are not included due to the memory constraint of a single WSE-2 chip.

Compared to T10, \sys achieves 160$\times$ speedup on average, up to 180$\times$, for short sequence generation tasks such as 4096 and 2048 input context lengths with 128 tokens output. For longer tasks, with input context lengths of 4096 and 2048 tokens and output lengths of 4096 and 2048 tokens, \sys achieves 36$\times$ on average and up to 48$\times$. Although T10 designs the compute-shift model that considers the memory constraints (M) and routing resource limits (R) of a PLMR device, it does not account for the cores interconnected by a mesh NoC. Thus, failing to address varying hop distances (L) and scale to millions of cores (P), highlighting the need for new system designs in massive-scale NUMA architectures.

Compared to Ladder, \sys achieves 625$\times$ speedup on average, up 677$\times$, for short sequence generation tasks such as 4096 and 2048 input context lengths with 128 tokens output. For longer sequence generation tasks, with input context lengths of 4096 and 2048 tokens and output lengths of 4096 and 2048 tokens, \sys achieves 312$\times$ on average and up to 342$\times$. That is because Ladder is designed for shared memory architecture and does not consider the characteristics of the PLMR device, resulting in failure in partitioning LLMs across millions of cores (P), incurring costly long-range NoC communication (L), failure in handling local memory constraints (M) and limited routing resources (R).

\begin{table}[t]
\centering
\resizebox{\linewidth}{!}{
\begin{tabular}{c|c|ccclccc}
\cline{1-5} \cline{7-9}
\multirow{2}{*}{Model} & \multirow{2}{*}{Methods} & \multicolumn{3}{c}{WSE-2 Cores} &  & \multicolumn{3}{c}{\# A100 GPUs (SGLang)} \\ \cline{3-5} \cline{7-9} 
 &  & \multicolumn{1}{c|}{480$\times$480} & \multicolumn{1}{c|}{600$\times$600} & 720$\times$720 &  & \multicolumn{1}{c|}{1} & \multicolumn{1}{c|}{8} & 2$\times$8 \\ \cline{1-5} \cline{7-9} 
\multirow{3}{*}{LLaMA3-8B} & \sys & \multicolumn{1}{c|}{20320.6} & \multicolumn{1}{c|}{25037.2} & 27686.5 &  & \multicolumn{1}{c|}{\multirow{3}{*}{13988.3}} & \multicolumn{1}{c|}{\multirow{3}{*}{17361.6}} & \multirow{3}{*}{13994.2} \\
 & T10 & \multicolumn{1}{c|}{175.0} & \multicolumn{1}{c|}{156.6} & 132.8 &  & \multicolumn{1}{c|}{} & \multicolumn{1}{c|}{} &  \\
 & Ladder & \multicolumn{1}{c|}{61.8} & \multicolumn{1}{c|}{42.3} & 31.3 &  & \multicolumn{1}{c|}{} & \multicolumn{1}{c|}{} &  \\ \cline{1-5} \cline{7-9} 
\multirow{3}{*}{LLaMA2-13B} & \sys & \multicolumn{1}{c|}{13685.1} & \multicolumn{1}{c|}{16854.2} & 17498.3 &  & \multicolumn{1}{c|}{\multirow{3}{*}{7805.1}} & \multicolumn{1}{c|}{\multirow{3}{*}{12287.1}} & \multirow{3}{*}{/\textsuperscript{\ref{missing_16GPU}}} \\
 & T10 & \multicolumn{1}{c|}{121.3} & \multicolumn{1}{c|}{100.6} & 81.3 &  & \multicolumn{1}{c|}{} & \multicolumn{1}{c|}{} &  \\
 & Ladder & \multicolumn{1}{c|}{47.3} & \multicolumn{1}{c|}{33.1} & 24.2 &  & \multicolumn{1}{c|}{} & \multicolumn{1}{c|}{} &  \\ \cline{1-5} \cline{7-9} 
\multirow{3}{*}{CodeLLaMA-34B} & \sys & \multicolumn{1}{c|}{5471.4} & \multicolumn{1}{c|}{7540.1} & 8526 &  & \multicolumn{1}{c|}{\multirow{3}{*}{5382.5}} & \multicolumn{1}{c|}{\multirow{3}{*}{7155.5}} & \multirow{3}{*}{6409.2} \\
 & T10 & \multicolumn{1}{c|}{49.1} & \multicolumn{1}{c|}{46.8} & 41.2 &  & \multicolumn{1}{c|}{} & \multicolumn{1}{c|}{} &  \\
 & Ladder & \multicolumn{1}{c|}{30.1} & \multicolumn{1}{c|}{23.1} & 17.7 &  & \multicolumn{1}{c|}{} & \multicolumn{1}{c|}{} &  \\ \cline{1-5} \cline{7-9} 
\multirow{3}{*}{QWen2-72B} & \sys & \multicolumn{1}{c|}{2785.2} & \multicolumn{1}{c|}{3775.5} & 4421.6 &  & \multicolumn{1}{c|}{\multirow{3}{*}{1677.3}} & \multicolumn{1}{c|}{\multirow{3}{*}{3803.8}} & \multirow{3}{*}{3750.5} \\
 & T10 & \multicolumn{1}{c|}{24.9} & \multicolumn{1}{c|}{23.5} & 21.5 &  & \multicolumn{1}{c|}{} & \multicolumn{1}{c|}{} &  \\
 & Ladder & \multicolumn{1}{c|}{16.8} & \multicolumn{1}{c|}{12.8} & 10.1 &  & \multicolumn{1}{c|}{} & \multicolumn{1}{c|}{} &  \\ \cline{1-5} \cline{7-9} 
\end{tabular}%
}
  \caption{Prefill Throughput per Request (TPR)}
  \vspace{-0.2cm}
\label{tab:prefill_throughput}
\end{table}%

\mypar{Prefill throughput} Table~\ref{tab:prefill_throughput} shows the prefill performance for an input sequence length of 4096, using core configurations from 480$\times$480 to 720$\times$720. For CodeLLaMA-34B and QWen2-72B, which exceed the memory capacity of WSE-2 and a single A100, we evaluate a subset of layers and scale the results proportionally due to their uniform layer structure.

\sys achieves significant speedups over T10 and Ladder by effectively addressing all PLMR properties, an average speedup of ~160$\times$ (up to 178$\times$) over T10 and ~270-450$\times$ over Ladder. As discussed in Section~\S\ref{sec:motivation}, GEMM is the primary bottleneck, and \gemm substantially enhances \sys's prefill performance, analyzed in detail in Section~\S\ref{sec:eval:meshgemm}.

Additionally, \sys scales throughput with increasing cores across all models. For instance, \sys achieves a 1.6$\times$ scaleup on QWen2-72B and a 1.4$\times$ scaleup on LLaMA3-8B when scaling from 480$\times$480 to 720$\times$720 cores. In contrast, T10 and Ladder fail to scale effectively, with throughput even declining as more cores are added. This is mainly because T10 and Ladder do not account for the spatial locality characteristic (L) of highly uniformly distributed memory, which can lead to frequent long-distance data reads across cores and increased communication overhead.

\begin{table}[t]
  \centering
  \resizebox{\linewidth}{!}{
    \begin{tabular}{c|c|ccclccc}
\cline{1-5} \cline{7-9}
\multirow{2}{*}{Model} & \multirow{2}{*}{Methods} & \multicolumn{3}{c}{WSE-2 Cores} &  & \multicolumn{3}{c}{\# A100 GPUs (SGLang)} \\ \cline{3-5} \cline{7-9} 
 &  & \multicolumn{1}{c|}{420$\times$420} & \multicolumn{1}{c|}{540$\times$540} & 660$\times$660 &  & \multicolumn{1}{c|}{1} & \multicolumn{1}{c|}{8} & 2$\times$8 \\ \cline{1-5} \cline{7-9} 
\multirow{3}{*}{LLaMA3-8B} & \sys & \multicolumn{1}{c|}{2699.9} & \multicolumn{1}{c|}{2501.5} & 2243.3 &  & \multicolumn{1}{c|}{\multirow{3}{*}{78.9}} & \multicolumn{1}{c|}{\multirow{3}{*}{260.4}} & \multirow{3}{*}{164.6} \\
 & T10 & \multicolumn{1}{c|}{418.3} & \multicolumn{1}{c|}{339.4} & 265.1 &  & \multicolumn{1}{c|}{} & \multicolumn{1}{c|}{} &  \\
 & Ladder & \multicolumn{1}{c|}{14.6} & \multicolumn{1}{c|}{13.1} & 11.4 &  & \multicolumn{1}{c|}{} & \multicolumn{1}{c|}{} &  \\ \cline{1-5} \cline{7-9} 
\multirow{3}{*}{LLaMA2-13B} & \sys & \multicolumn{1}{c|}{2039.2} & \multicolumn{1}{c|}{1899.4} & 1739.8 &  & \multicolumn{1}{c|}{\multirow{3}{*}{48.7}} & \multicolumn{1}{c|}{\multirow{3}{*}{175.8}} & \multirow{3}{*}{/\textsuperscript{\ref{missing_16GPU}}} \\
 & T10 & \multicolumn{1}{c|}{341.8} & \multicolumn{1}{c|}{270.8} & 233.7 &  & \multicolumn{1}{c|}{} & \multicolumn{1}{c|}{} &  \\
 & Ladder & \multicolumn{1}{c|}{11.0} & \multicolumn{1}{c|}{9.9} & 9.0 &  & \multicolumn{1}{c|}{} & \multicolumn{1}{c|}{} &  \\ \cline{1-5} \cline{7-9} 
\multirow{3}{*}{CodeLLaMA-34B} & \sys & \multicolumn{1}{c|}{1450.8} & \multicolumn{1}{c|}{1407.7} & 1359.2 &  & \multicolumn{1}{c|}{\multirow{3}{*}{26.1}} & \multicolumn{1}{c|}{\multirow{3}{*}{100.4}} & \multirow{3}{*}{84.5} \\
 & T10 & \multicolumn{1}{c|}{278.2} & \multicolumn{1}{c|}{222.4} & 193.1 &  & \multicolumn{1}{c|}{} & \multicolumn{1}{c|}{} &  \\
 & Ladder & \multicolumn{1}{c|}{6.1} & \multicolumn{1}{c|}{6.2} & 5.8 &  & \multicolumn{1}{c|}{} & \multicolumn{1}{c|}{} &  \\ \cline{1-5} \cline{7-9} 
\multirow{3}{*}{QWen2-72B} & \sys & \multicolumn{1}{c|}{839.7} & \multicolumn{1}{c|}{824.3} & 787.1 &  & \multicolumn{1}{c|}{\multirow{3}{*}{10.6}} & \multicolumn{1}{c|}{\multirow{3}{*}{51.2}} & \multirow{3}{*}{48.7} \\
 & T10 & \multicolumn{1}{c|}{168.5} & \multicolumn{1}{c|}{133.0} & 114.6 &  & \multicolumn{1}{c|}{} & \multicolumn{1}{c|}{} &  \\
 & Ladder & \multicolumn{1}{c|}{3.2} & \multicolumn{1}{c|}{3.3} & 3.4 &  & \multicolumn{1}{c|}{} & \multicolumn{1}{c|}{} &  \\ \cline{1-5} \cline{7-9} 
\end{tabular}%
    }
  \caption{Decode Throughput per Request (TPR)}
  \vspace{-0.2cm}
\label{tab:decode_throughput}
\end{table}%

\mypar{Decode throughput} Table~\ref{tab:decode_throughput} shows decode throughput for core configurations from 420$\times$420 to 660$\times$660. For CodeLLaMA-34B and QWen2-72B decode benchmark on WSE-2 and single A100, we evaluate a subset of layers and scale the results.

By addressing all PLMR properties, \sys achieves an average speedup of 5.7$\times$ (up to 6.5$\times$) over T10 and 217$\times$ (up to 260$\times$) over Ladder.

Compared to the 160$\times$ speedup \sys achieves over T10 during prefill, it offers only about a 6.5$\times$ speedup during decode. However, \sys maintains a consistent 200-500$\times$ advantage over Ladder in both prefill and decode stages. This is mainly because prefill involves dist-GEMM computations, which require each data element to traverse cores along the same row or column sequentially. In contrast, the dist-GEMV allreduce operation in decode only requires traversal across cores in the same row or column, without enforcing a strict access order. While T10’s compute-shift paradigm does not consider the spatial locality of cores on a mesh NoC, it still accounts for distributed memory, which allows performance gains when memory (communication) accesses are order-independent. In contrast, Ladder is entirely designed for a shared memory architecture, leading to severe communication overhead regardless of whether memory access order is required.

\mypar{Comparison with SGLang} As shown in Table~\ref{tab:e2e_table},\ref{tab:prefill_throughput} and \ref{tab:decode_throughput}, \sys demonstrates clear advantages over SGLang + A100 GPU clusters in prefill, decode tasks, and also the end-to-end performance across models ranging from 8B to 72B. More detailed comparisons with GPUs are discussed in Section~\S\ref{sec:eval:gpu}.

It is worth noting that T10 and Ladder fail to leverage the powerful hardware capabilities of the WSE-2 when compared to SGLang. For example, in prefill tasks and for Ladder in decode tasks, their performance is even worse than that of SGLang running on a single A100. This highlights that designing high-performance systems and algorithms on wafer-scale chips must carefully consider the PLMR model; otherwise, not only will the massive compute potential of the chip go underutilized, but performance may even degrade due to the highly uniform distributed memory resulting from the large-scale mesh NoC.

\vspace{-0.2cm}
\subsection{\gemm}\label{sec:eval:meshgemm}
\vspace{-0.2cm}

\begin{figure}
    \centering
    \includegraphics[width=1\linewidth]{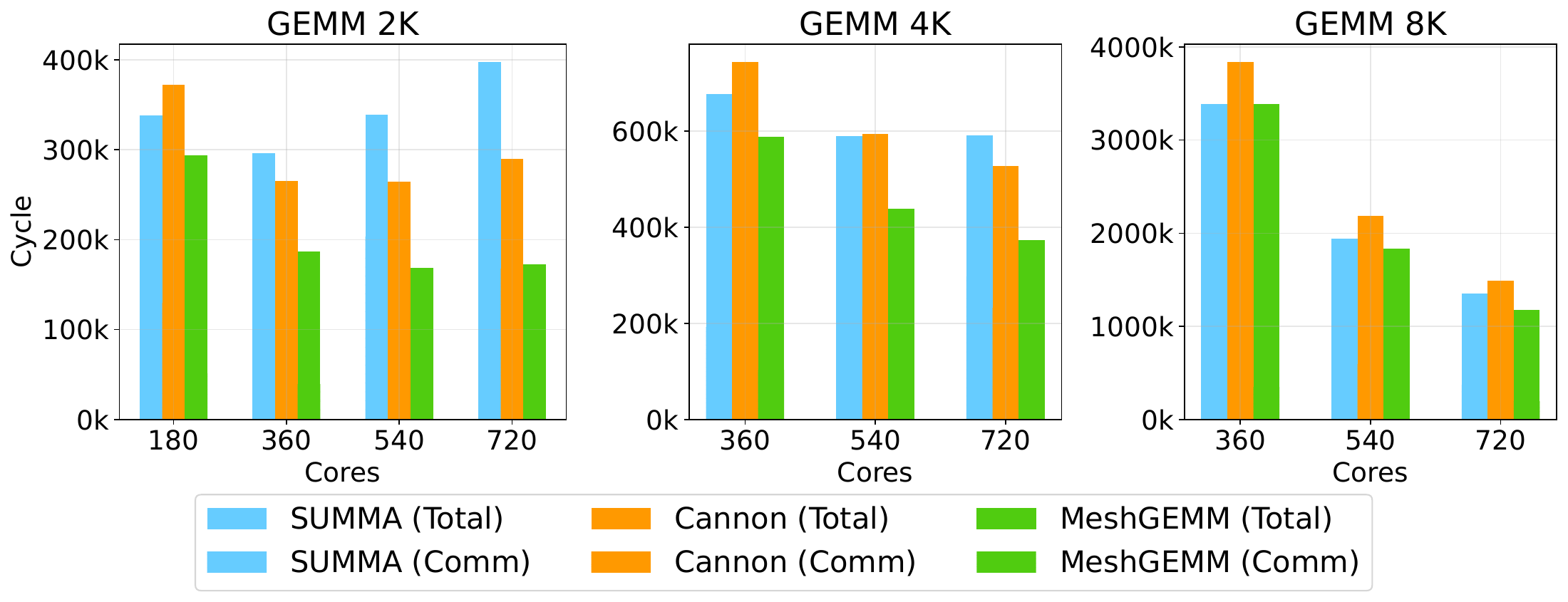}
    \caption{\gemm vs. SUMMA \& Cannon}
    \vspace{-0.5cm}
    \label{fig:eval:meshgemm}
\end{figure}

We compare \gemm with Cannon~\cite{cannon} and SUMMA~\cite{summa} on WSE-2 across different core scales and matrix sizes.

\mypar{Scaling core count} 
Figure~\ref{fig:eval:meshgemm} shows that across matrices of different sizes, \gemm can achieve the lowest latency than SUMMA and Cannon by scaling up the number of cores. Compared to the pipeline broadcast in SUMMA and head-to-tail transmission in Cannon, interleave transmission in \gemm can minimize per-step communication overhead and, as much as possible, overlap communication with computation during large-scale fine-grained parallel execution. It demonstrates stronger scalability, maintaining over 70\% computational efficiency even near the hardware limit. In contrast, SUMMA and Cannon exhibit poor scalability, with computational efficiency falling below 50\% with 720$\times$720 cores, primarily due to the communication overhead in SUMMA and Cannon, increasing with the parallelism scale.

Additionally, increasing the number of cores does not always bring performance gains, especially for small matrix GEMM tasks. For example, in GEMM 2K, when scaling from 360$\times$360 to 720$\times$720 cores, the end-to-end latency of SUMMA and Cannon increases instead of decreasing. This is because the per-core computation cost drops sublinearly as cores increase beyond a threshold (due to fixed overheads like function calls and logic checks). At the same time, communication overhead grows linearly, as seen from the shaded areas in Figure~\ref{fig:eval:meshgemm}, leading to worse overall performance. In contrast, \gemm only shows a slight communication overhead increase at 720$\times$720 cores and maintains stable end-to-end latency. This is because its interleave communication bounds the per-step communication cost to a constant, independent of core count (the total communication overhead grows only because the number of steps increases in all dist-GEMM algorithms). Thus, \gemm can achieve a better overlapping between communication and computation even under large-scale fine-grained parallelism (each core is only responsible for a small amount of data).

\mypar{Scaling matrix size} We also evaluate \gemm with larger matrix sizes, transforming GEMM into a more computationally intensive operation. At large scales, though the cost of communication becomes less significant, \gemm maintains its scalability and outperforms SUMMA and Cannon by a wide margin, reducing total cycles by around 17\%.

An interesting observation in Figure~\ref{fig:eval:meshgemm} is that for GEMM 8K, communication cycles decrease as core count increases. This occurs because when processing large data volumes, communication is bandwidth-bound rather than latency-bound. Increasing the number of cores not only boosts aggregate compute power to reduce computation overhead but also increases aggregate bandwidth to lower overall communication cost.

\vspace{-0.2cm}
\subsection{\gemv}
\vspace{-0.2cm}

We evaluate \gemv and the default GEMV implementation on Cerebras (pipeline allreduce) across various core scales and matrix shapes.

\begin{figure}
    \centering
    \includegraphics[width=1\linewidth]{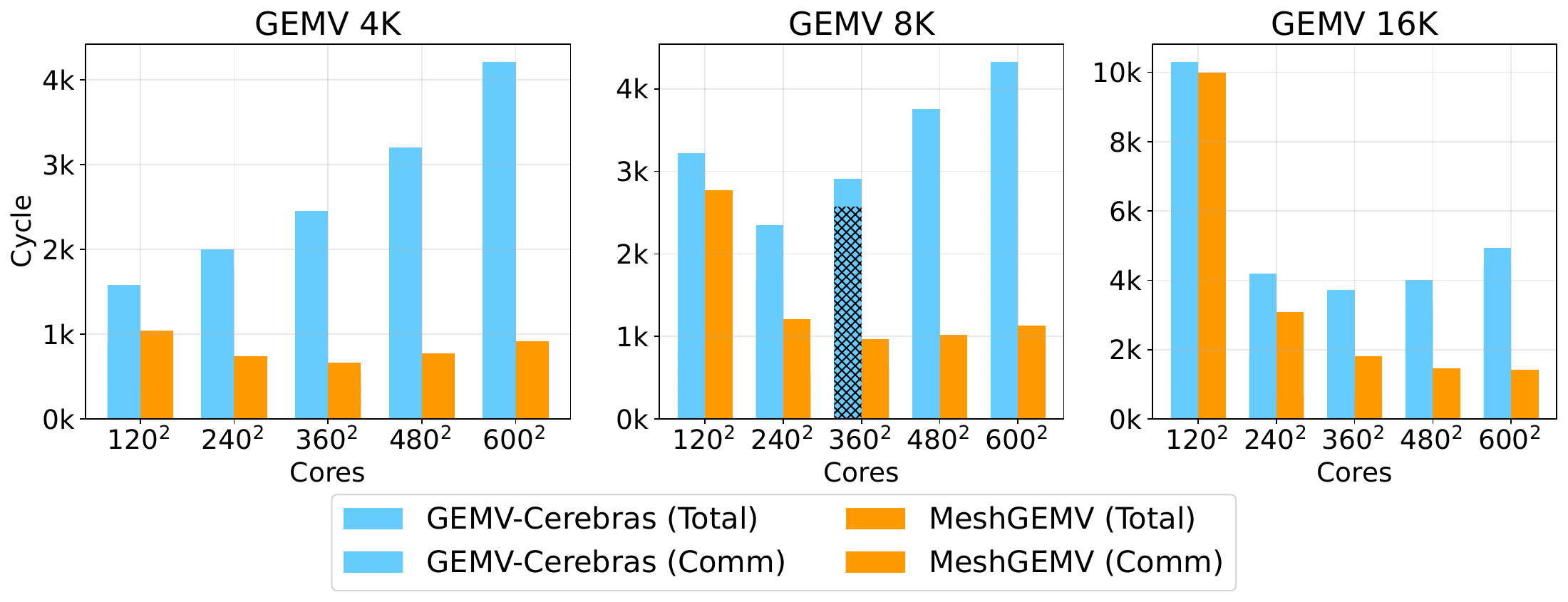}
    \caption{\gemv vs. GEMV-Cerebras}
    \vspace{-0.3cm}
    \label{fig:eval:meshgemv}
\end{figure}

\mypar{Communication latency bottleneck} Figure~\ref{fig:eval:meshgemv} shows that communication becomes the primary bottleneck for dist-GEMV computations on WSE-2, especially when the parallelism scale is large relative to the computation per core, where communication overhead can dominate 90\% of the total. However, \gemv significantly reduces communication overhead compared to the Cerebras baseline GEMV, achieving about 4.6$\times$ higher end-to-end performance. This is because large-scale dist-GEMV requires long-distance allreduce communication, and \gemv uses K-tree Allreduce to maximize parallelism in the allreduce process, minimizing communication and computation costs along the critical path.

\mypar{Scaling core count and matrix size} We also evaluate \gemv with larger matrix sizes. For GEMV 8K and 16K compared to 2K, the baseline shows a trend that the end-to-end overhead first decreases and then increases. The initial decrease comes from increased core count, where the higher aggregate compute power and bandwidth reduce computation and communication costs. However, as the core count grows, the communication latency of allreduce eventually overtakes computation and bandwidth as the dominant bottleneck. In contrast, \gemv's inflection point appears later, which means that for the same data size, \gemv's communication overhead grows much more slowly with increasing core count compared to the baseline, demonstrating better scalability. This enables more flexible on-chip mapping policies, such as using more cores to store additional LLM parameters to meet memory limitations (M) without introducing significant extra overhead.

\vspace{-0.3cm}
\subsection{Shift-based KV cache management}\label{sec:eval:kvcache}
\vspace{-0.2cm}

\begin{table}[t]
  \centering
  \resizebox{0.8
  \linewidth}{!}{
    \begin{tabular}{l|c|c}
    \hline
    Model & LLaMA3-8B & LLaMA2-13B \\
          \hline
    Concat-based (PagedAttention) & 382   & 16 \\
    \hline
    Shift-based (\sys) & 137548 & 6168 \\
    \hline
    \end{tabular}%
    }
  \caption{Maximum decode output length}
  \label{tab:eval:kvcache}%
\end{table}%

We also compare the shift-based KV cache management with the concat-based KV cache management in PagedAttention. We evaluate KV cache capacity on LLaMA3-8B and LLaMA2-13B using the same settings as the end-to-end inference evaluation in Section~\S\ref{sec:eval:e2e}. Table~\ref{tab:eval:kvcache} shows that \sys's shift-based KV cache management supports 360$\times$ and 385$\times$ more tokens than the concat-based method for LLaMA3-8B and LLaMA2-13B, respectively. This improvement results from balanced core utilization and the resolution of skewed data issues achieved by the shift-based approach.

\vspace{-0.2cm}
\subsection{Comparison with GPUs}\label{sec:eval:gpu}
\vspace{-0.2cm}

\begin{table}[t]
  \centering
  \resizebox{\linewidth}{!}{
    \begin{tabular}{c|ccc|ccc}
\hline
GEMV & \multicolumn{3}{c|}{[1,16K]$\times$[16K,16K]} & \multicolumn{3}{c}{[1,32K]$\times$[32K,32K]} \\ \hline
\multirow{2}{*}{\begin{tabular}[c]{@{}c@{}}TP in SGLang(A100)\\ Time (ms)\end{tabular}} & 1 GPU & 8 GPUs & 2$\times$8 GPUs & 1 GPU & 8 GPUs & 2$\times$8 GPUs \\ \cline{2-7} 
 & 0.336 & 0.253 & 0.340 & 1.231 & 0.341 & 0.339 \\ \hline
\begin{tabular}[c]{@{}c@{}}\gemv(WSE-2)\\ Time (ms)\end{tabular} &  & 0.0012 &  &  & 0.00203 &  \\ \hline
\begin{tabular}[c]{@{}c@{}}A100/WSE-2 Energy Ratio\end{tabular} & \textbf{7.47} & \textbf{44.97} & \textbf{120.88} & \textbf{16.17} & \textbf{35.83} & \textbf{71.25} \\ \hline
\end{tabular}%
    }
  \caption{Comparing \gemv(WSE-2) with TP in SGLang(A100) of GEMV latency and energy.}
  \vspace{-0.3cm}
  \label{tab:mv_gpu}%
\end{table}%

We compare \sys against the state-of-the-art LLM inference system on GPUs using Cerebras WSE-2 and NVIDIA A100. To compare against the H100 fairly, we would need access to the WSE-3, which is a 5nm manufacturer but unavailable to us.

\mypar{GEMV} We compared \gemv with the GEMV parallelization strategy that follows SGLang's \cite{sglang} multi-GPU tensor parallelism, while the computation on each GPU is accelerated using the cuBLAS library. Shown by Table~\ref{tab:mv_gpu}, compared to a single A100 GPU, \gemv outperforms by 280-606$\times$ with different matrix sizes, showcasing the advantages of providing substantial memory bandwidth through wafer-scale devices. This also translates to 7.5-16$\times$ greater energy efficiency, reflecting the benefits of wafer-based connections (connecting on-chip memory) over PCB-based ones (connecting off-chip HBM) in GPUs.

Distributed GEMV on multi-GPUs shows limited scalability: a single A100 yields the highest energy efficiency per FLOP, and performance improves only 1.32$\times$ from one to eight GPUs, then degrades at sixteen. This inefficiency stems from memory intensity and communication overhead over NVLink and IB. In contrast, \gemv exploits WSE-2's low-latency NoC to scale across hundreds of thousands of cores, achieving 166-210$\times$ higher performance and 45-70$\times$ better energy efficiency at peak compared to the best GPU cluster results. Moreover, GEMV exhibits similar performance for 16K and 32K tasks on two machines with 16 GPUs, mainly because communication overhead dominates the computation in distributed GEMV across two nodes.

Despite these advantages, \gemv does not achieve the theoretical 7,000$\times$ improvement. Profiling identifies three contributing factors: (i)~WSE-2 cores, still in their second generation, cannot fully overlap memory access and computation; (ii)~edge cores are underutilized; and (iii)~NoC long-range communication overhead persists, despite \gemv mitigating it effectively. We anticipate these gaps will continue to narrow as wafer-scale accelerators mature.

\begin{table}[t]
  \centering
  \resizebox{\linewidth}{!}{
    \begin{tabular}{c|ccc|cc}
\hline
Prefill (4K CTX) & \multicolumn{3}{c|}{LLaMA3-8B} & \multicolumn{2}{c}{LLaMA2-13B} \\ \hline
\multirow{2}{*}{\begin{tabular}[c]{@{}c@{}}SGLang(A100) TPR\end{tabular}} & 1 GPU & 8 GPUs & 2$\times$8 GPUs & 1 GPU & 8 GPUs \\ \cline{2-6} 
 & 13988 & 17361 & 13994 & 7805 & 12287 \\ \hline
\begin{tabular}[c]{@{}c@{}}\sys WSE-2 TPR\end{tabular} &  & 27686 &  & \multicolumn{2}{c}{17498} \\ \hline
A100/WSE-2 Energy Ratio & \textbf{0.05} & \textbf{0.34} & \textbf{0.84} & \textbf{0.06} & \textbf{0.30} \\ \hline
\end{tabular}%
    }
  \caption{Comparing \sys(WSE-2) with SGLang(A100) in prefill throughput and energy.}
  \vspace{-0.3cm}
  \label{tab:llmoc_gpu_prefill}%
\end{table}%

\begin{table}[t]
  \centering
  \resizebox{\linewidth}{!}{
    \begin{tabular}{c|ccc|cc}
\hline
Decode (4K CTX) & \multicolumn{3}{c|}{LLaMA3-8B} & \multicolumn{2}{c}{LLaMA2-13B} \\ \hline
\multirow{2}{*}{\begin{tabular}[c]{@{}c@{}}SGLang(A100) TPR\end{tabular}} & 1 GPU & 8 GPUs & 2$\times$8 GPUs & 1 GPU & 8 GPUs \\ \cline{2-6} 
 & 78 & 260 & 164 & 48 & 175 \\ \hline
\begin{tabular}[c]{@{}c@{}}\sys WSE-2 TPR\end{tabular} &  & 2700 &  & \multicolumn{2}{c}{2039} \\ \hline
A100/WSE-2 Energy Ratio & \textbf{0.92} & \textbf{2.22} & \textbf{7.02} & \textbf{1.13} & \textbf{2.49} \\ \hline
\end{tabular}%
    }
  \caption{Comparing \sys(WSE-2) with SGLang(A100) in decode throughput and energy.}
  \vspace{-0.3cm}
  \label{tab:llmoc_gpu_decode}%
\end{table}%

\mypar{LLM inference} The maximum TPR achievable on GPU clusters is significantly lower than on WSE-2. We compare \sys with SGLang on an A100 multi-GPUs cluster and find that \sys delivers a 6-20$\times$ higher TPR across a range of input/output lengths and model sizes from 8B to 72B, with the gap widening for longer outputs and larger models (Table~\ref{tab:e2e_table}). As shown in Tables~\ref{tab:llmoc_gpu_prefill} and \ref{tab:llmoc_gpu_decode}, scaling SGLang from 1 to 8 GPUs yields only 1.2-1.6$\times$ prefill and 3.3-3.6$\times$ decode speedups, far below ideal linear scaling. Performance further degrades when scaling to 16 GPUs, due to inter-node communication bottlenecks. As a result, peak TPR for dense models is typically reached within a single 8-GPU node, where \sys still outperforms SGLang up to 20$\times$. For use cases demanding high per-request throughput, wafer-scale accelerators are substantially more capable than conventional xPU-based systems.

In terms of energy efficiency, though a WSE-2 chip has ~47$\times$ the area and ~37$\times$ the power and cost of an A100, \sys still achieves a 2-2.5$\times$ energy efficiency advantage at SGLang's optimal multi-GPU result, owing to GPUs' nonlinear scaling on decode. This advantage is especially valuable for long-output scenarios like test-time scaling, where serving cost is critical. However, compared to GEMV, \sys's decode energy efficiency advantage is reduced due to: (i) limited local SRAM (48KB) on WSE-2 cores, which hinders efficient tensor parallelism and necessitates pipeline parallelism, causing up to 5$\times$ underutilization; and (ii) GPU-optimized LLaMA models with narrow layers that constrain layer placement and worsen utilization on WSE-2.

\vspace{-0.3cm}
\section{Implementation Detail and Future Direction}

We discuss the current limitations of \sys and wafer-scale accelerators and envision their future solutions:

\mypar{Hardware architecture} The performance of \sys is currently constrained by execution bubbles caused by the need for pipeline parallelism. Increasing a core's local memory by 5-6$\times$ could eliminate the need for pipeline parallelism, enabling full tensor parallelism, as on vLLM and SGLang. Wafer-scale chip designers are already moving in this direction. Cerebras WSE-3 retains the same NoC configuration but improves per-core efficiency and local memory, while Tesla’s Dojo incorporates 1MB of per-core memory.

\mypar{Memory-to-Compute Ratio} LLM decoding demands a near 1:1 memory-to-compute ratio. However, GPUs like A100 have limited on-chip SRAM, forcing frequent off-chip memory access and yielding a poor ratio of 1:312 (FP16). Multi-GPU setups exacerbate this due to heavy inter-GPU communication. 
In contrast, Cerebras WSE-2 maps most model weights onto on-chip memory via a low-latency NoC, achieving near-ideal locality and approaching a 1:1 ratio. To fully realize this balance on mesh-based NoCs, adherence to the PLMR model is essential, enabling \sys to outperform GPU-based systems by orders of magnitude in TPR.

\mypar{Handle reliability issues} Currently, Cerebras WSE-2/3 handles faults by hardware, only exposing healthy cores (organized in a mesh) to software, with no explicit handling required at the software level. Moreover, redundant cores and links are built in at fabrication, and the on-chip SoC dynamically remaps and reroutes around defects at runtime, ensuring minimal performance impact at a low redundancy cost. Meanwhile, wafer-scale chip makers recently reported a 93\% functional wafer area, higher than 70–80\% in commercial GPUs~\cite{cerebras_defect_tolerance}, due to the smaller area per core design. Over two years of WSE-2 deployment, we have observed high reliability, confirming the effectiveness of these fault-tolerance mechanisms in real-world use.

\mypar{Various model architecture} \sys is also beneficial for MoE as it shares key operators with dense LLMs, including \gemm, \gemv, and shift-based KV cache management. The main difference is the all-to-all communication between attention and expert layers, which we implement using WSE-2's NoC multi-cast operations. Further optimizations for sparse models, such as offloading and sparse attention, are among our future research.

\mypar{Beyond Cerebras WSE} While evaluated on Cerebras WSE, the PLMR model generalizes to emerging mesh-like architectures such as Tesla Dojo, which also feature hundreds of thousands of cores with local memory and constrained NoC routing. Variants like 2D torus or hybrid mesh-switch topologies also conform to PLMR. Our design for \gemm and \gemv targets worst-case 2D mesh and remains competitive across such platforms. Beyond on-chip meshes, chip-to-chip mesh interconnects, as seen in Tenstorrent’s core- and card-level meshes, also align well with PLMR. Looking ahead, advancements in wafer-scale integration, such as TSMC’s projected 40$\times$ density increase by 2027, further reinforce the long-term applicability of our approach.

\vspace{-0.2cm}
\section{Related Work}
\vspace{-0.2cm}

\mypar{Deep learning frameworks and compilers}
Current deep learning frameworks and compilers, such as PyTorch, TensorFlow, and XLA~\cite{pytorch, tensorflow, xla,flextensor, tvm, ansor, roller, rammer, welder, ladder}, are designed for shared memory architectures and use a tile-based “load-compute-store” computation model. While effective for shared memory, this model ignores the unique characteristics of PLMR devices, making it inefficient for wafer-scale AI chips. LLM frameworks such as vLLM and TensorRT-LLM~\cite{vllm,tensorrt-llm} have emerged to support modern LLMs but rely on frameworks and compilers designed for shared memory architectures (e.g., PyTorch~\cite{pytorch}), inheriting similar limitations on wafer-scale chips.

\mypar{Distributed GPU and TPU systems} The on-chip distributed memory architecture could theoretically be treated as a distributed LLM system, as studied in prior works~\cite{alpa, alpaserver, vllm, google2023efficiently, tensorrt-llm, loongserve}. However, such systems, designed for GPU and TPU pods (up to thousands of nodes), rely on more capable routers and lack local memory constraints, making them misaligned with the PLMR model. These approaches are complementary to our focus on on-chip scaling.

\mypar{Systolic array} Systolic array architectures~\cite{resa}, used in AI accelerators such as Amazon Trainium and Google TPU, focus on the design of small cores rather than larger wafer-scale accelerators. With limited processing elements (usually up to hundreds) in a core, they are not PLMR devices but complement \sys. For example, a Cerebras WSE core could employ a systolic array to accelerate local GEMM operations.

\mypar{Dataflow architectures}
Prior research has explored computation on dataflow architectures that account for inter-core connections. TENET~\cite{tenet} maps computation spatially and temporally to connected cores in a dataflow pattern. DISTAL~\cite{distal} enables scheduling over distributed clusters using a dataflow approach. SambaNova~\cite{sambanova} combines model and pipeline parallelism for DNN execution. However, none of these works scale computation to wafer-scale chips.

\mypar{Wafer-scale allreduce} Recent research~\cite{wafer-reduce} has investigated wafer-scale allreduce, but a single allreduce cannot fully parallelize GEMV or support full LLM inference as achieved by \sys. Additionally, this prior work is a specific instance of the K-tree allreduce proposed in \sys.

\vspace{-0.3cm}
\section{Conclusion}

We envision this paper as a foundational step in exploring the potential of wafer-scale computing for LLMs. The simple yet effective PLMR model has revealed significant opportunities, guiding the development of the first wafer-scale LLM parallelism solution and scalable GEMM and GEMV algorithms for wafer-scale accelerators. Despite the limitations of the current software stack for wafer-scale devices, our approach achieves orders-of-magnitude improvements in both performance and energy efficiency. We hope this work inspires greater focus on wafer-scale computing and advances the path toward a more sustainable future for AI.

\vspace{-0.3cm}
\section{Acknowledgments}

We sincerely thank our shepherd, Marco Canini, and the OSDI reviewers for their insightful feedback that greatly improved this paper. We are also grateful to Yuqing Xia (Microsoft) for her valuable input during the development of this work. 

We acknowledge the hardware resources provided by the Edinburgh International Data Facility (EIDF) and the Edinburgh Parallel Computing Centre (EPCC), particularly their support in granting access to the Cerebras WSE systems. We thank Nick Johnson for coordinating access and Mark Parsons for supporting the hardware acquisition and our project.

Finally, we appreciate the generous assistance of Leighton Wilson and Mathias Jacquelin from Cerebras for their patience and detailed responses to our many questions regarding CSL and the Cerebras WSE.

\clearpage
\bibliographystyle{plain}
\bibliography{main}


\end{document}